\documentclass[pdflatex,sn-mathphys,smallcondensed]{sn-jnl}

\jyear{2023}%
\usepackage{algpseudocode}

\usepackage{caption}
\usepackage{graphicx}
\usepackage{comment}
\usepackage{natbib}
\theoremstyle{thmstyleone}%

\usepackage{bm,graphicx}
\usepackage{amsmath}
\usepackage{fullwidth}
\usepackage{subcaption}
\usepackage[algo2e]{algorithm2e}

\usepackage{array}
\newcolumntype{P}[1]{>{\centering\arraybackslash}p{#1}}

\usepackage[inline, shortlabels]{enumitem}

\theoremstyle{thmstyletwo}%

\theoremstyle{thmstylethree}%

\begin{document}

\title{ADSumm: Annotated Ground-truth Summary Datasets for Disaster Tweet Summarization}


\author*[1]{Piyush Kumar Garg}\email{piyush\_2021cs05@iitp.ac.in}

\author[2]{Roshni Chakraborty}\email{roshni.chakraborty@ut.ee}

\author[3]{Sourav Kumar Dandapat}\email{sourav@iitp.ac.in}

\affil[1]{\orgdiv{Department of Computer Science and Engineering}, \orgname{Indian Institute of Technology Patna}, \orgaddress{\state{Bihar}, \country{India}}}

\affil[2]{\orgdiv{Institute of Computer Science}, \orgname{University of Tartu}, \orgaddress{\country{Estonia}}}

\abstract{
Online social media platforms, such as Twitter, provide valuable information during disaster events. Existing tweet disaster summarization approaches provide a summary of these events to aid government agencies, humanitarian organizations, etc., to ensure effective disaster response. In the literature, there are two types of approaches for disaster summarization, namely, supervised and unsupervised approaches. Although supervised approaches are typically more effective, they necessitate a sizable number of disaster event summaries for testing and training. However, there is a lack of good number of disaster summary datasets for training and evaluation. This motivates us to add more datasets to make supervised learning approaches more efficient. In this paper, we present \textit{ADSumm}, which adds annotated ground-truth summaries for eight disaster events which consist of both natural and man-made disaster events belonging to seven different countries. Our experimental analysis shows that the newly added datasets improve the performance of the supervised summarization approaches by $8-28$\% in terms of ROUGE-N F1-score. Moreover, in newly annotated dataset, we have added a \textit{category label} for each input tweet which helps to ensure good coverage from different categories in summary. Additionally, we have added two other features \textit{relevance label} and \textit{key-phrase}, which provide information about the quality of a tweet and explanation about the inclusion of the tweet into summary, respectively. For ground-truth summary creation, we provide the annotation procedure adapted in detail, which has not been described in existing literature. Experimental analysis shows the quality of ground-truth summary is very good with \textit{Coverage}, \textit{Relevance} and \textit{Diversity}. 
}


\maketitle

\section{Introduction} \label{s:intro}

\par During disasters, eyewitnesses and bystanders share information through social networks, such as Twitter~\cite{imran2016twitter}. Several prior research works highlight that tweets are a valuable source of information during disasters~\cite{basu2019extracting, priya2019should, dutt2019utilizing, ghosh2018exploitation}. This information consists of updates about casualties, infrastructure damage, information on missing people, urgent needs of resources and supplies, etc. Many disaster response organizations, government agencies, volunteers, and NGOs rely on this valuable information to plan and launch relief operations immediately~\cite{castillo2016big}. However, tweets are inherently short and often contain grammatical errors, abbreviations, and informal language. These characteristics make it highly challenging to identify relevant information from them. Additionally, the huge number of tweets makes it highly challenging for any human being and government organizations to identify relevant information manually~\cite{alam2020descriptive, Alam2021humaid}. 

        
        
    

\par To handle these challenges, several research works~\cite{garg2023ontodsumm, dusart2023tssubert, rudra2019summarizing, dutta2018ensemble} have been proposed for disaster tweet summarization that generate an automated summary of the disaster tweets. Existing disaster tweet summarization approaches are broadly categorized into supervised and unsupervised approaches. However, supervised approaches are more efficient than unsupervised approaches. Efficiency of these automated supervised summarization approaches depend heavily on the availability of sizable ground-truth summaries for training. Furthermore, there is a high variance in information among disasters on the basis of the type and location of a disaster~\cite{garg2023ontodsumm}. Therefore, the non-availability of ground-truth summaries for disasters of different locations and types affects the development of a robust and efficient summarization approach. Although Dutta et al.~\cite{dutta2018ensemble} and Rudra et al.~\cite{rudra2018extracting} have provided the ground-truth summary of six datasets (shown in Table~\ref{table:existdata}) which has been of huge help to the research community, there are around $50$ different disaster datasets~\cite{olteanu2015expect, alam2018crisismmd, Alam2021humaid} which do not have any ground-truth summary. The addition of more ground-truth summaries of different disasters will surely improve this situation which motivates us to add more ground truth summaries.

\par In existing literature, we have not found any systematic approach for ground truth creation. Existing approaches completely rely on the wisdom and knowledge of annotators, where a flat set of input tweets is provided to annotators to generate a summary. In this paper, to come up with a systematic approach, we mimic the important steps followed by automatic summarization approaches~\cite{garg2023ontodsumm, rudra2015extracting}, which could be described as 1) identification of the category of each tweet, 2) understanding of the importance of each category, and 3) selection of important tweets from each category for summary. However, most of these above-mentioned steps are subjective. While automated approaches take important decisions for summary creation based on some well-defined approach and parameters, annotators need to make these decisions based on their wisdom. Following the existing literature~\cite{garg2023portrait}, we have selected three annotators for this task. This work presents \textit{ADSumm}, which refers to the ground-truth summaries that have been created by three annotators for a total of eight distinct disaster events, each occurring in different locations and of varying types. The inclusion of these datasets would greatly benefit the research community. Furthermore, the incorporation of the recently introduced datasets enhances the efficacy of supervised summarization approaches by $8-28$\% in relation to the ROUGE-N F1-score, as elaborated in the corresponding Subsection~\ref{s:impact}.

\par  In addition to the ground-truth summaries of eight datasets, we provide annotation of three supplementary features that are not yet present in the public datasets. The aforementioned features encompass \textit{category labels}, \textit{key-phrases}, and \textit{relevance labels}. The term \textit{category label} pertains to a specific classification assigned to a tweet that falls within the context of a disaster. These categories may include \textit{Infrastructure Damage, Volunteer Operations, Affected Population, Impact, Prayer}, and others. The \textit{relevance label} assigned to a tweet indicates its significance in relation to the associated disaster occurrence, classifying it as \textit{high}, \textit{medium}, or \textit{low}. The \textit{key-phrase} elucidates the potential rationale underlying the significance of a tweet. For instance, a tweet, ``Tropical storm Hagupit kills at least 21 in Philippines'' can be classified under the \textit{category label} \textit{Affected Population}. The tweet contains a \textit{key-phrase} \textit{storm Hagupit kills at least 21 in Philippines}, which is \textit{high}ly relevant to the Hagupit Typhoon\footnote{\url{https://en.wikipedia.org/wiki/Typhoon\_Hagupit\_(2014)}} disaster event. We offer the \textit{category label, key-phrase, and relevance labels} for each tweet in eight distinct disaster datasets, which might be of great assistance to the academic research community. 

\par  In this paper, we also present comprehensive analyses on the utility of the additional dataset features in assessing the effectiveness for various advanced natural language processing (NLP) tasks. These tasks can be broadly divided into three categories 1) disaster tweet classification, 2) development of a robust summarization algorithm, and 3) evaluation of the quality of the disaster summary. Feature \textit{category label} provides ground truth for the tweet category, which can be used to assess the classification accuracy of any classification approach. As feature \textit{key-phrase} helps in ranking tweets within a category and calculate diversity in the summary,  \textit{key-phrase} plays an important role in robust summarization algorithm development. Furthermore, the feature \textit{relevance label} immensely helps to assess the summary quality as it provides quantitative understanding of the importance of each tweet. We also conduct a comparative analysis of $16$ existing state-of-the-art summarization approaches to evaluate their efficacy in summarizing eight datasets related to disaster events. This evaluation serves as a reference for benchmarking purposes. The datasets were utilized as additional ground-truth summaries as described in~\cite{garg2023ontodsumm}. 

\begin{table*}
    \caption{Table shows the details of available six disaster datasets, which include dataset name, number of tweets, summary length, country, continent, and disaster type.}
    \label{table:existdata}
    \resizebox{\textwidth}{!}{\begin{tabular} {lccccc}
        \hline
        {\bf Dataset name} & {\bf Number of tweets} & {\bf Summary length} & {\bf Country} & {\bf Continent} & {\bf Disaster type}\\ \hline
        
        \textit{Sandy Hook Elementary School Shooting}  & 2080 & 36 tweets & United States  & North America & Man-made   \\ 
        \textit{Uttrakhand Flood}                       & 2069 & 34 tweets & India          & Asia & Natural  \\ 
        \textit{Hagupit Typhoon}                        & 1461 & 41 tweets & Philippines    & Asia & Natural  \\ 
        \textit{Hyderabad Blast}                        & 1413 & 33 tweets & India          & Asia & Man-made \\ 
        \textit{Harda Twin Train Derailment}            & 4171 & 250 words & India          & Asia & Man-made \\ 
        \textit{Nepal Earthquake}                       & 5000 & 250 words & Nepal          & Asia & Natural  \\ \hline
    \end{tabular}}
\end{table*}

\par The rest of the paper is organized as follows. We discuss related works in Section~\ref{s:rworks}. In Section~\ref{s:data}, we provide the details of the datasets and discuss the ground-truth preparation details in Section~\ref{s:dataprep}. In Section~\ref{s:results}, we discuss results where we provide qualitative and quantitative analysis results of the annotated ground-truth summary in Subsection~\ref{s:quantcomp}. We provide an inter-annotator agreement between all the annotators to check the consistency for ground-truth generation in Subsection~\ref{s:ianoag}. We compare summaries generated by the supervised approaches to assess the importance of additional training set for summary generation in Subsection~\ref{s:impact}. We also provide a case study where we show the use cases of the dataset's additional features in the different NLP-related tasks in Subsection~\ref{s:utilifeat}. We discuss the experiment details and results of the performance comparison of the various existing state-of-the-art summarization approaches in Subsection~\ref{s:exp}. Finally, we conclude the paper in Section~\ref{s:con}.

\section{Related Works} \label{s:rworks}
\par Summarization provides a brief summary of input text, emphasizing its essential aspects and the key information~\cite{nazari2019survey}. Text summarization approaches have been proposed for various domains, such as legal texts summarization~\cite{jain2023bayesian}, news summarization~\cite{duan2019across}, timeline summarization~\cite{ansah2019graph}, tweet summarization~\cite{chakraborty2019tweet,chakraborty2017network}, etc. Similarly, there are several disaster specific summarization approaches~\cite{roy2020classification, dutta2019summarizing, Garg2022Entropy} which could be categorized into content and context based~\cite{rudra2015extracting, li2021twitter}, graph-based approaches~\cite{dutta2015graph} and deep learning based approaches~\cite{de2019time}. 
    
\par For example, content-based disaster tweet summarization approaches \cite{rudra2016summarizing, rudra2018extracting, rudra2019summarizing} explore the importance of keywords to generate a summary which might not ensure coverage in summary. Recent deep learning-based approaches have proposed techniques, such as Dusart et al.~\cite{dusart2023tssubert} integrated the importance of each word by frequency into Bidirectional Encoder Representations from Transformers (BERT)~\cite{devlin2018bert}, Garg et al.~\cite{garg2024ikdsumm} integrated the \textit{key-phrase} of each tweet into BERT, and Li et al.~\cite{li2021twitter} capture the inter-tweet similarity through a Graph Convolution Neural (GCN) network to generate the summary. However, these approaches require extensive training and therefore, more number of disaster events with annotated summaries. Graph-based summarization approaches~\cite{dutta2018ensemble, dutta2019community, rudra2019summarizing} initially generate a graph of tweets where the nodes are tweets and the edges represent the similarity between a pair of tweets followed by identification of groups of similar tweets using different approaches, such as communities~\cite{fortunato2010community}, connected components~\cite{gazit1991optimal}, and $k$-clique clustering~\cite{kim2014tweet}. Finally, they select the representative tweets from each group to create a summary. Although these existing approaches of identifying groups are highly effective in news event-based tweet summarization~\cite{chakraborty2019tweet,chakraborty2017network}, they fail for disaster events where there is a high vocabulary overlap across different groups. Therefore, Rudra et al.~\cite{rudra2016summarizing, rudra2018identifying} and Garg et al.~\cite{garg2023ontodsumm} initially identify the category of a tweet and then select representative tweets from each category into the summary. While Rudra et al.~\cite{rudra2016summarizing, rudra2018identifying} use Artificial Intelligence for Disaster Response (AIDR)~\cite{imran2014aidr}, which requires human intervention for each new disaster in real-time, Garg et al.~\cite{garg2023ontodsumm} proposed an un-supervised ontology-based category identification approach that does not require any human intervention for a disaster event. However, the vocabulary and importance of the category differ on the basis of the type and location of the disaster. Therefore, deep learning, graph-based, and category-based approaches require ground-truth summaries of different types and locations to learn the inherent differences. 

\par Existing research works, such as Rudra et al.~\cite{rudra2018extracting} and Dutta et al.~\cite{dutta2018ensemble} have provided the ground-truth summary of six disaster events (shown in Table~\ref{table:existdata}) belonging to both the type of disaster (man-made and natural) and four different countries. However, these datasets are not good enough to train a supervised summarization approach. Therefore, these approaches could not reach the required performances, as discussed in Section~\ref{s:impact}. Although, the existing datasets did not provide any supplementary features, such as \textit{key-phrases}, \textit{category labels}, and \textit{relevance labels}. These features, along with the datasets, could be further used for evaluating different NLP-related tasks such as disaster tweet classification, developing a robust summarization algorithm, and evaluating the quality of disaster summary. The non-availability of ground-truth summaries along with features for different locations and types affects the development of a robust summarization algorithm and quality checking which is suitable for disaster events irrespective of the variance in disaster data. Therefore, in this paper, we present the annotated ground-truth summaries along with the three different features with the dataset for eight different disaster events belonging to natural and man-made disasters and seven more different countries.

\section{Datasets} \label{s:data}
\par In this Section, we discuss the datasets for which we prepare the ground-truth summaries. We show the details for $D_1$-$D_8$ datasets in Table~\ref{table:Edatasets}.  

\begin{table*}
    \caption{Table shows the details of $D_1$-$D_8$ datasets, which include dataset number, dataset name, year, number of tweets, summary length, country, continent, and disaster type.}
    \label{table:Edatasets}
    \centering
    \resizebox{\textwidth}{!}{\begin{tabular} {clcccccc}
        \hline
        {\bf Num} & {\bf Name} & {\bf Year} & {\bf Number of tweets} & {\bf Summary length} & {\bf Country} & {\bf Continent} & {\bf Disaster type}\\\hline  
        
        $D_1$ & \textit{Los Angeles International Airport Shooting} & 2013 & 1409 & 40 tweets & United States & North America & Man-made \\ 
        $D_2$ & \textit{Hurricane Matthew}                          & 2016 & 1654 & 40 tweets & Haiti & North America & Natural\\ 
        $D_3$ & \textit{Puebla Mexico Earthquake}                   & 2017 & 2015 & 40 tweets & Mexico & North America & Natural \\ 
        $D_4$ & \textit{Pakistan Earthquake}                        & 2019 & 1958 & 40 tweets & Pakistan & Asia & Natural\\ 
        $D_5$ & \textit{Midwestern U.S. Floods}                  & 2019 & 1880 & 40 tweets & United States & North America & Natural\\ 
        $D_6$ & \textit{Kaikoura Earthquake}                     & 2016 & 2195 & 40 tweets & New Zealand & Oceania & Natural \\ 
        $D_7$ & \textit{Cyclone Pam}                             & 2015 & 1508 & 40 tweets & Vanuatu & Oceania & Natural\\ 
        $D_8$ & \textit{Canada Wildfires}                        & 2016 & 2242 & 40 tweets & Canada & North America & Natural  \\ \hline
    \end{tabular}}
\end{table*}

\begin{enumerate}
    
    
    
    

    \item \textit{$D_1$}: This dataset is created by~\cite{olteanu2015expect} from a terrorist attack on the \textit{Los Angeles International Airport Shooting}\footnote{\url{https://en.wikipedia.org/wiki/2013\_Los\_Angeles\_International\_Airport\_shooting}} on November, $2013$ in which $1$ person was killed and more than $15$ people were injured. 
    
    \item \textit{$D_2$}: This dataset is created by~\cite{Alam2021humaid} from the \textit{Hurricane Matthew}\footnote{\url{https://en.wikipedia.org/wiki/Hurricane\_Matthew}} on October, $2016$ in which around $603$ people were killed, around $128$ people were missing and the estimated damage were around \$$2.8$ billion USD. 
    
    \item \textit{$D_3$}: This dataset is created by~\cite{Alam2021humaid} from the \textit{Puebla Mexico Earthquake}\footnote{\url{https://en.wikipedia.org/wiki/2017\_Puebla\_earthquake}} on September, $2017$ in which $370$ people were dead and more than $6000$ people were injured. 
    
    \item \textit{$D_4$}: This dataset is created by~\cite{Alam2021humaid} from the \textit{Pakistan Earthquake}\footnote{\url{https://en.wikipedia.org/wiki/2019\_Kashmir\_earthquake}} on September, $2019$ in which around $40$ people were killed, $850$ people were injured, and around $319$ houses were damaged. 
    
    \item \textit{$D_5$}: This dataset is created by~\cite{Alam2021humaid} from the \textit{Midwestern U.S. Floods}\footnote{\url{https://en.wikipedia.org/wiki/2019\_Midwestern\_U.S.\_floods}} in which around $14$ million people were affected, and damage were around $2.9$ billion USD.

    \item \textit{$D_6$}: This dataset is created by~\cite{Alam2021humaid} from the \textit{Kaikoura Earthquake}\footnote{\url{https://en.wikipedia.org/wiki/2016\_Kaikoura\_earthquake}} on November, $2016$ in which $2$ people were died, and around $57$ were injured. 
    
    \item \textit{$D_7$}: This dataset is created by~\cite{imran2016twitter} from the \textit{Cyclone Pam}\footnote{\url{https://en.wikipedia.org/wiki/Cyclone\_Pam}} in March, $2015$ in which around $15$ people were died and around $3300$ people were displaced.
    
    \item \textit{$D_8$}: This dataset is created by~\cite{Alam2021humaid} from the \textit{Canada Wildfires}\footnote{\url{https://en.wikipedia.org/wiki/2016\_Fort\_McMurray\_wildfire}} in May, $2016$ in which $1,456,810$ acres were burned, $3,244$ buildings were destroyed, and damage were around $9.9$ billion C\$.
\end{enumerate}

\noindent\textbf{Dataset Pre-processing}\\ 
We perform standard pre-processing steps, like, lemmatization, conversion to lowercase, removal of noisy keywords, white spaces, and punctuation marks. Additionally, we consider only tweet text and subsequently we remove the Twitter-specific keywords like hashtags, URLs, usernames, and emoticons~\cite{arachie2020unsupervised}. To reduce the redundancy, we further remove retweets followed by duplicate tweets. Lastly, we follow Alam et al.~\cite{alam2018crisismmd} to remove noise, i.e., any word which is of length less than three characters. We show some examples of tweets for $D_3$ and $D_6$ in Table~\ref{table:Tweets}.

\begin{table*}[ht]
    \caption{Table shows some examples of tweets text of two disaster events, such as \textit{Puebla Mexico Earthquake} $(D_3)$ and \textit{Kaikoura Earthquake} $(D_6)$.}
    \label{table:Tweets}
    \centering 
    \begin{tabular}{p{0.15\linewidth}p{0.75\linewidth}} \hline
    \textbf{Event} & \textbf{Tweet text} \\ \hline
            
                                & 7.1 magnitude earthquake in Mexico kills 226 people. \\ 
                                & RT @latimes: 2,000 historic buildings in Mexico have been damaged by the earthquake. \\ 
    \textit{Puebla Mexico Earthquake}      & Hundreds of rescue personnel and citizens in Mexico City have banded together to rescue earthquake survivors. \\ 
                                & RT @Kevinwoo91: My prayers are continuing to be with everyone who has been affected by the earthquake in Mexico \#PrayForMexico \\ 
                                & Venezuela Delivers 10.4 Tons of Aid to Earthquake-Ravaged Mexico. \\ \hline
                                
                                & RT @NZcivildefence: 14 staff from Urban Search \&amp; Rescue have been deployed to \#Wellington to help assess buildings. Another teams on stand \\ 
                                & \#New Zealand PM John Key says 2 people killed in earthquake; sending military helicopter to Kaikoura - Reuters \\ 
    \textit{Kaikoura Earthquake}& RT @nytimes: An earthquake measuring 7.9 hit New Zealand, triggering 3 large aftershocks and at least 3 tsunami waves \\
                                & Reports of damage to at least 25 buildings in Wellington so far \#eqnz \#wellington \\ 
                                & Praying for \#NZ - still a Tsunami threat from the 7.5 mag \#eqnz. May Allah keep everyone safe. Pls follow instructions from @NZcivildefence \\ \hline
    \end{tabular}
\end{table*}

\begin{figure*}
    \centering \includegraphics[width=\textwidth] {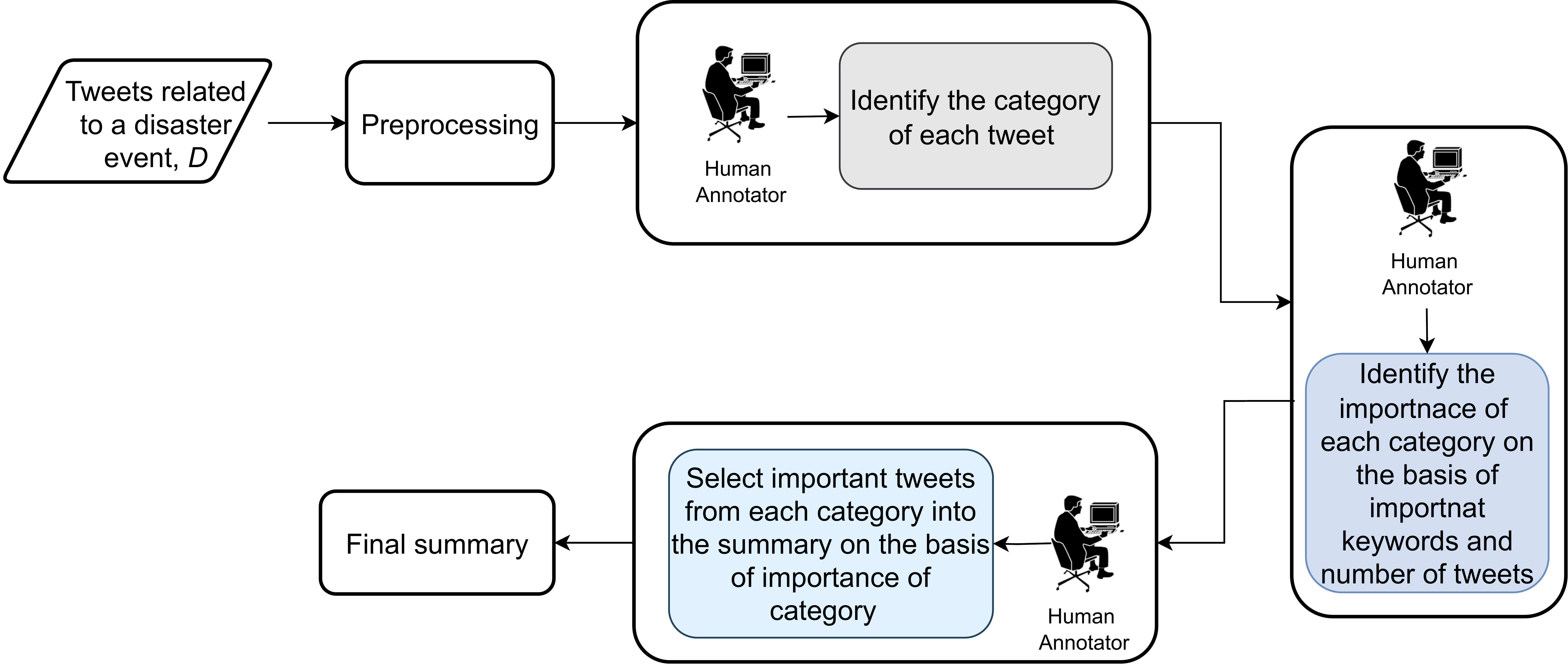}
    \caption{An overview of the proposed systematic annotation procedure is shown.}
    \label{figure:flowchart}
\end{figure*}

\section{Ground-truth Preparation}\label{s:dataprep}
\par In this Section, we discuss the annotation procedure followed by annotators to generate a ground-truth summary of a disaster event. In the existing literature~\cite{dutta2018ensemble, rudra2015extracting, rudra2018identifying}, we find that annotators are provided a flat set of tweets of a disaster event and prepare the summary based on their knowledge and wisdom. Annotators manually assess the importance of each tweet with respect to the disaster event and then decide whether it should be part of the summary based on intuition given all the tweets related to a disaster event. These approaches mainly depend on the understanding of the annotators related to that disaster event. Therefore, there is a need for a systematic approach to ground-truth summary creation from a flat collection of tweets. Additionally, ground-truth summary generation is a subjective task, so we can not depend on only one annotator for the summary, and we require at least three annotators for their individual summaries~\cite{poddar2022caves, rudra2018extracting, dutta2018ensemble}. Additionally, the ground-truth generation mainly depends on the annotator's wisdom related to a disaster event, and therefore, there is a need to ensure the quality and consistency of the annotators before selecting them for the ground-truth summary generation task. For this, we follow Garg et al.~\cite{garg2023portrait} where we select three annotators on the basis of a \textit{Quality Assessment Test}. We refer to these annotators as $A_1$, $A_2$, and $A_3$ in the rest of the paper. We next discuss the details of the systematic approach followed by the annotators to generate the final ground-truth summary.

\par To develop a systematic approach, we imitate the important steps followed by the existing automatic summarization approaches~\cite{garg2023ontodsumm, rudra2015extracting}, which could be as follows: 1) identification of the category of each tweet, 2) identification of the importance of each category on the basis of tweets' important keywords and the number of tweets, and 3) selection of important tweets from each category in summary on the basis of the importance of that category. However, the majority of the aforementioned steps involve subjective judgment, and therefore, for each dataset, we request all three annotators to go through all the tweets given a dataset and prepare a ground-truth summary with the given summary length by following these steps manually. We follow Dutta et al.~\cite{dutta2018ensemble} and Rudra et al.~\cite{rudra2015extracting} to decide the summary length as $40$ tweets. For the first step of ground-truth creation, which is category identification of each tweet, prior works~\cite{rudra2019summarizing, imran2016twitter, imran2015towards} have highlighted that the tweets of a disaster event belong to different categories. A ground-truth summary of a disaster should provide \textit{Coverage} all important aspects\footnote{In this paper, we refer aspects by tweet category.}. However, the dataset does not comprise the category of the tweet inherently. To handle this, we share a publicly available ontology for disasters, namely, \textit{Empathi}~\cite{gaur2019empathi}, which comprises of $70$ categories. A set of keywords are also provided associated with these categories. We request the annotators to initially segregate all the tweets into these different ontology categories. Based on the keyword similarity (similarity between keywords of a category and keywords of a tweet), a tweet is assigned a category label by annotators. As a next step, annotators identify the importance of each category on the basis of both the number of tweets and the important keywords of the tweets in each category. Further, depending on the importance of a category, an annotator decides how many tweets to be selected from a category on the basis of his/her judgment of the importance of that category with respect to the disaster. An annotator can even consider a category to be not important enough to be covered in the final summary. Finally, annotators rank tweets within each category to select a specific number of tweets from a category in the summary. Annotators also take care about diversity during the selection of tweets from a category. An overview of this systematic annotation procedure is shown in Figure~\ref{figure:flowchart}.

\section{Results and Discussions} \label{s:results}
\par In this Section, we evaluate the quality (by means of qualitative and quantitative analysis) of annotated ground-truth summaries using three metrics, namely \textit{Coverage}, \textit{Relevance}, and \textit{Diversity}. Further, to check the consistency among the annotators, we provide an inter-annotator agreement between all three annotators for two different tasks: 1) category assignment and 2) ground-truth summary. We further provide representation through the word cloud to visualise the annotated ground-truth summaries of a disaster event. To assess the importance of additional training set for the supervised approach of summary generation, we compare summaries generated by the supervised approach for two scenarios: 1) when trained with existing datasets and 2) when trained with existing datasets along with additional datasets. We also provide a case study where we show the use cases of the dataset's additional features in the different NLP-related tasks. Finally, to provide benchmark data, we compare and evaluate the existing state-of-the-art summarization approaches on the annotated ground-truth summaries for $D_1-D_8$ datasets.

\subsection{Qualitative and Quantitative Evaluation of a Ground-truth Summary} \label{s:qqeval}
\par In this Subsection, we evaluate the quality of the generated ground-truth summary by means of quantitative and qualitative analysis based on \textit{Coverage}, \textit{Relevance}, and \textit{Diversity}. We discuss each of the evaluation measures next.

\subsubsection*{\textbf{Qualitative Evaluation}} 
\par We follow Garg et al.~\cite{garg2023portrait} for qualitative evaluation of the summary quality. A summary is considered to be of acceptable quality if it has high \textit{Coverage}, \textit{Relevance}, and \textit{Diversity}. We ask three meta-annotators\footnote{The meta-annotators are graduate students who belong to the age group of $20-30$, have good knowledge of English and are not a part of this project.} to score the ground-truth summaries on a scale of $1$ (worst score) to $5$ (best score) for each summary on the basis of its fulfilment of \textit{Coverage}, \textit{Relevance} and \textit{Diversity} as \textit{Coverage Score}, \textit{Relevance Score}, and \textit{Diversity Score}, respectively. For each summary, we get all the above-mentioned scores from all three meta-annotators. To combine scores from different annotators, we aggregate all the scores and take an average of that. For example, if \textit{Coverage Score} of a summary from three meta-annotators are x, y, and z, respectively, then the \textit{Aggregate Coverage Score} can be computed as (x+y+z)/3. In a similar way, we have also computed \textit{Aggregate Relevance Score} and \textit{Aggregate Diversity Score}. We show our observations in Table~\ref{table:qualicomp}, which indicates that the ground-truth summaries are of high quality in terms of \textit{Coverage}, \textit{Relevance}, and \textit{Diversity}. 

\begin{table}[htbp]
    \caption{Table shows the \textit{Aggregate Coverage Score}, \textit{Aggregate Relevance Score}, and \textit{Aggregate Diversity Score} of the annotated ground-truth summaries of all the three annotators for $D_1-D_8$ datasets.}
    \label{table:qualicomp}
    \centering
    \resizebox{0.95\linewidth}{!}{\begin{tabular} {cccc||cccc}
        \hline
        {\bf Dataset} & {\bf Aggregate} & {\bf Aggregate} & {\bf Aggregate} & {\bf Dataset} & {\bf Aggregate} & {\bf Aggregate} & {\bf Aggregate}\\
                      & {\bf Coverage}  & {\bf Relevance} & {\bf Diversity} & & {\bf Coverage}  & {\bf Relevance} & {\bf Diversity}\\
                      & {\bf Score}     & {\bf Score}     & {\bf Score} & & {\bf Score}     & {\bf Score}     & {\bf Score}\\ \hline 
        
        $D_1$       & 3.94 & 4.19 & 3.86 & $D_5$       & 3.69 & 3.44 & 4.42 \\         
        $D_2$       & 3.78 & 4.22 & 3.38 & $D_6$       & 4.22 & 4.19 & 3.86 \\  
        $D_3$       & 4.33 & 3.31 & 3.94 & $D_7$       & 3.67 & 4.44 & 3.67 \\  
        $D_4$       & 4.05 & 3.47 & 3.75 & $D_8$       & 4.00 & 3.67 & 4.25 \\ \hline  
    \end{tabular}}
\end{table}

\subsubsection*{\textbf{Quantitative Evaluation}} \label{s:quantcomp}
\par In this Subsection, we follow Garg et al.~\cite{garg2023portrait} to asses the annotated summaries through \textit{Coverage}, \textit{Relevance} and \textit{Diversity}, quantitatively.

\par \textbf{Coverage :} Coverage is a well-accepted metric for assessing summary quality and provides a measure of how many important aspects are included in the summary. In this paper, we realize different aspects by means of category~\cite{garg2023portrait}. In order to compute this, we identify the category coverage in the annotator's annotated ground-truth summaries. We utilize the categories identified by the annotators during ground-truth preparation discussed in Section~\ref{s:dataprep}. We show the number of categories in the annotated ground-truth summaries of three annotators and in a dataset for $D_1-D_8$ datasets in Table~\ref{table:catcoverage}. Similar to the qualitative assessment, for quantitative analysis, we also compute the \textit{Aggregate Coverage Score} for each ground-truth summary. Our observations indicate that the \textit{Aggregate Coverage} of all three annotators is in the range of $51.85-77.78$\% across the datasets. The $D_6$ dataset has the highest \textit{Aggregate Coverage}  ($77.78$\%), and the $D_3$ dataset has the lowest \textit{Aggregate Coverage} ($51.85$\%). Furthermore, we observe that for the categories that are not captured in the summary, we found that both the number of tweets and the importance of those categories with respect to the disaster event are very low. Hence, we can say that the generated ground-truth summaries ensure good \textit{Coverage} (covering important aspects/categories). 

\begin{table}[htbp]
    \caption{We show the number of categories in a dataset and the ground-truth summaries and Aggregate Coverage (in \%) of all three annotators for $D_1-D_8$ datasets. (Note: \# represents a number in this table.)}
    \label{table:catcoverage}
    \smallskip
    \centering
    \resizebox{\linewidth}{!}{\begin{tabular} {cccccc||cccccc}
        \hline
        {\bf Dataset} & \bf{\# of categories}  & \multicolumn{3}{c}{\bf \# of categories covered in } & \bf Aggregate & {\bf Dataset} & \bf{\# of categories}  & \multicolumn{3}{c}{\bf \# of categories covered in } & \bf Aggregate\\ 
                      & \bf{covered in a}   & \multicolumn{3}{c}{\bf ground-truth summary of} & \bf Coverage & & \bf{covered in a}   & \multicolumn{3}{c}{\bf ground-truth summary of} & \bf Coverage \\
                      &  \bf{dataset} &  {\bf $A_1$} & {\bf $A_2$} & {\bf $A_3$} & \bf in \% & &  \bf{dataset} &  {\bf $A_1$} & {\bf $A_2$} & {\bf $A_3$} & \bf in \%\\ \hline 
        
        $D_1$   & 9  & 5 & 5 & 7 & 65.32 & $D_5$   & 8  & 5 & 7 & 6 & 75.00 \\  
        $D_2$   & 9  & 6 & 6 & 7 & 70.37 & $D_6$   & 9  & 7 & 7 & 7 & 77.78 \\ 
        $D_3$   & 9  & 4 & 5 & 5 & 51.85 & $D_7$   & 9  & 7 & 7 & 7 & 77.78 \\ 
        $D_4$   & 10 & 6 & 8 & 8 & 73.33 & $D_8$   & 9  & 5 & 5 & 6 & 59.26 \\ \bottomrule 
    \end{tabular}}
\end{table}

\par \textbf{Relevance :} A summary has a limited capacity to capture and preserve the comprehensive details and nuances present in a dataset. Therefore, one measure of assessing the summary is \textit{Relevance}, which quantifies the inclusion of highly relevant tweets in the summary. In order to compute this, we ask all the meta-annotators to assign a label to all the tweets based on tweet's importance with respect to the disaster, as \textit{high}, \textit{medium}, and \textit{low}. Finally, label of a tweet is decided based on the majority vote of the three annotators. If we fail to decide based on a majority vote, we ask another annotator to decide the final label. However, for our datasets, we could assign all the labels on the basis of a majority vote. We show the percentage of each \textit{relevance label} for ground-truth summaries for $D_1-D_8$ datasets in Table~\ref{table:informlabel}. Our observations indicate that $57.50-75.00\%$ of the ground-truth summary tweets contain \textit{high} \textit{relevance label} across the datasets. Therefore, based on these observations, we can say that the generated ground-truth summaries comprise of significantly high-relevance tweets.

\begin{table*}[htbp]
    \centering
    \caption{Table shows the percentage number of tweets of each \textit{relevance label}, such as \textit{high}, \textit{medium}, and \textit{low} for the annotated ground-truth summaries of all the three annotators for $D_1-D_8$ datasets. (Note: we release the final relevance label based on the majority vote for each dataset.)}
    \label{table:informlabel}
    \resizebox{0.9\textwidth}{!}{\begin{tabular}{cccc|ccc|ccc}
        \hline
        \textbf{Dataset} & \multicolumn{3}{c}{$\bf A_1$} & \multicolumn{3}{c}{$\bf A_2$} & \multicolumn{3}{c}{$\bf A_3$} \\ \cline{2-10}
        & \textbf{High} & \textbf{Medium} & \textbf{Low} & \textbf{High} & \textbf{Medium} & \textbf{Low} & \textbf{High} & \textbf{Medium} & \textbf{Low} \\ \hline 
        $D_1$   & 60.00\% & 20.00\% & 20.00\% & 65.00\% & 17.50\% & 17.50\% & 60.00\% & 17.50\% & 22.50\% \\ 
        $D_2$   & 57.50\% & 15.00\% & 27.50\% & 65.00\% & 12.50\% & 22.50\% & 60.00\% & 20.00\% & 20.00\% \\ 
        $D_3$   & 65.00\% & 15.00\% & 20.00\% & 62.50\% & 15.00\% & 22.50\% & 70.00\% & 7.50\%  & 22.50\% \\ 
        $D_4$   & 60.00\% & 12.50\% & 27.50\% & 60.00\% & 12.50\% & 27.50\% & 57.50\% & 12.50\% & 30.00\% \\ 
        $D_5$   & 65.00\% & 22.50\% & 12.50\% & 65.00\% & 10.00\% & 25.00\% & 60.00\% & 17.50\% & 22.50\% \\ 
        $D_6$   & 60.00\% & 32.50\% & 7.50\%  & 65.00\% & 20.00\% & 15.00\% & 65.00\% & 22.50\% & 12.50\% \\ 
        $D_7$   & 60.00\% & 22.50\% & 15.00\% & 75.00\% & 15.00\% & 10.00\% & 57.50\% & 20.00\% & 22.50\% \\ 
        $D_8$   & 60.00\% & 20.00\% & 20.00\% & 65.00\% & 15.00\% & 20.00\% & 62.50\% & 15.00\% & 22.50\% \\ \hline
    \end{tabular}}
\end{table*}

\par \textbf{Diversity :} \textit{Diversity} is a well-accepted metric to evaluate the summary quality in terms of novel information present in summary~\cite{garg2023ontodsumm}. We follow Garg et al.~\cite{garg2023portrait} and Nguyen et al.~\cite{nguyen2022towards, nguyen2022rationale} to calculate \textit{Diversity} of a summary, which measures the mean \textit{Diversity} between every pair of tweets in the summary using tweets \textit{key-phrases}. For this purpose, we also requested a meta-annotator to provide \textit{key-phrases} of each tweet containing sufficient justification of its importance as well as information regarding the topic of the tweet. The meta-annotator follows the guideline where he/she initially identifies all the numeric, locations, and then identifies all the keywords relevant to disaster, and finally, he/she selects a \textit{key-phrase} as a continuous set of words from a tweet. For instance, let us consider a tweet, \textit{Tropical storm Hagupit kills at least 21 in Philippines}. As per the guideline, the following words will be marked- \textit{storm}, \textit{Hagupit}, \textit{kills}, \textit{21}, and \textit{Philippines}. Therefore the \textit{key-phrase} of the tweet which includes all the important keywords identified is \textit{storm Hagupit kills at least 21 in Philippines}. We calculate the \textit{Aggregate Diversity Score} over all the three annotators for eight disaster datasets. Our results, as shown in Table~\ref{table:diveval1}, indicate that the annotated ground-truth summaries contain $0.3885-0.6595$ \textit{Aggregate Diversity Score} across the disasters. The $D_2$ dataset has the highest \textit{Aggregate Diversity Score} ($0.6595$), and the $D_6$ dataset has the lowest \textit{Aggregate Diversity Score} ($0.3885$). Therefore, the above experiment implies that the annotated ground-truth summaries ensure good diversity in the summary tweets.

\begin{table}[ht!]
    \caption{Table shows the \textit{Aggregate Diversity Score} of the ground-truth summaries generated by the three annotators for $D_1-D_8$ disaster datasets.}
    \label{table:diveval1}
    \centering
    \resizebox{0.7\textwidth}{!}{\begin{tabular} {cc||cc}
        \hline
        {\bf Dataset} & {\bf Aggregate Diversity} & {\bf Dataset} & {\bf Aggregate Diversity} \\ 
                        & {\bf Score } & & {\bf Score } \\ \hline
        $D_1$   & 0.5404 & $D_5$ & 0.4342 \\
        $D_2$   & 0.6595 & $D_6$ & 0.3885 \\  
        $D_3$   & 0.5175 & $D_7$ & 0.4267 \\ 
        $D_4$   & 0.5206 & $D_8$ & 0.4202 \\ \hline
    \end{tabular}}
\end{table}

\par \textbf{Quality Comparison of existing datasets with our dataset : } To understand the significance of the quantitative quality of the annotated ground-truth summaries, we compare these summaries with the existing available datasets through \textit{Coverage}, \textit{Relevance}, and \textit{Diversity}. For comparison, we consider four available disaster datasets provided by Dutta et al.~\cite{dutta2018ensemble} i.e., \textit{Sandy Hook Elementary School Shooting}, \textit{Uttrakhand Flood}, \textit{Hagupit Typhoon}, and \textit{Hyderabad Blast} (ground-truth summaries annotated by three annotators\footnote{We refer these three annotators as $E_1$, $E_2$, and $E_3$, hereby} along with a tweet set). For above-mentioned datasets, we requested our three annotators for \textit{category label} annotations, three meta-annotators for \textit{relevance label} annotations, and one meta-annotator for \textit{key-phrase} annotations. Then, we use generated ground truth summaries of those datasets as published by the respective authors~\cite{dutta2018ensemble} to compute \textit{Coverage}, \textit{Relevance} and \textit{Diversity}. 

For \textit{Covergae}, we show the number of categories in the annotated ground-truth summaries of three annotators and in a dataset in Table~\ref{table:catcoverage1}. Our observations indicate that the \textit{Aggregate Coverage} of all three annotators is in the range of $51.85-74.09$\% across the datasets. For \textit{Relevance}, we show the percentage of each \textit{relevance label} for ground-truth summaries for these datasets in Table~\ref{table:catcoverage1}. Our observation indicates that $52.78-69.69$\% of the ground-truth summary tweets of existing datasets contain \textit{high relevance label} across the datasets. Similarly, we show the \textit{Aggregate Diversity Score} over all the three annotators for above four disaster datasets in Table~\ref{table:catcoverage1}. Our observations indicate that the ground-truth summary of existing datasets contains $0.4327-0.7034$ \textit{Aggregate Diversity Score} across the disasters. Therefore, we conclude that the quality of the proposed annotated ground-truth summaries is in the range of the quality of ground-truth summaries of existing datasets in terms of \textit{Coverage}, \textit{Relevance}, and \textit{Diversity}.

\begin{table}[htbp]
    \caption{We show the number of categories in a dataset and the ground-truth summaries and Aggregate Coverage (in \%) of all the three annotators, the percentage number of tweets of each \textit{relevance label}, such as \textit{high}, \textit{medium}, and \textit{low} for the annotated ground-truth summaries of all the three annotators, and \textit{Aggregate Diversity Score} of the ground-truth summaries generated by the three annotators for four existing disaster datasets, i.e., \textit{Sandy Hook Elementary School Shooting} (SHShoot), \textit{Uttrakhand Flood} (UFlood), \textit{Hagupit Typhoon} (HTyphoon), and \textit{Hyderabad Blast} (HBlast). (Note: \# represents a number in this table.)}
    \label{table:catcoverage1}
    \smallskip
    \centering
    \resizebox{\linewidth}{!}{\begin{tabular} {lccccc|ccc|ccc|ccc|c}
        \hline
        {\bf Dataset} & \bf{\# of categories}  & \multicolumn{3}{c}{\bf \# of categories covered in } & \bf Aggregate & \multicolumn{9}{c|}{\bf Relevance label} & {\bf Aggregate}\\ 
                      & \bf{covered in a}   & \multicolumn{3}{c}{\bf ground-truth summary of} & \bf Coverage & \multicolumn{3}{c}{$\bf E_1$} & \multicolumn{3}{c}{$\bf E_2$} & \multicolumn{3}{c|}{$\bf E_3$} & {\bf Diversity }\\
                      &  \bf{dataset} &  {\bf $E_1$} & {\bf $E_2$} & {\bf $E_3$} & \bf in \% & \textbf{High} & \textbf{Medium} & \textbf{Low} & \textbf{High} & \textbf{Medium} & \textbf{Low} & \textbf{High} & \textbf{Medium} & \textbf{Low} &{\bf Score }
                      \\ \hline 
        
        SHShoot   & 9  & 5 & 6 & 4 & 55.55 & 63.89 & 16.67 & 19.44 & 52.78 & 36.11 & 11.11 & 63.89 & 16.67 & 19.44  & 0.4327 \\  
        UFlood   & 9  & 4 & 5 & 5 & 51.85 & 64.71 & 11.76 & 23.53 & 52.94 & 26.47 & 20.59 & 58.88 & 20.59 & 20.59 & 0.6319 \\
        HTyphoon   & 9  & 6 & 7 & 7 & 74.07 & 60.98 & 24.39 & 14.63 & 58.53 & 21.95 & 19.51 & 63.41 & 17.07 & 19.51 & 0.7034 \\
        HBlast   & 9 & 5 & 5 & 6 & 59.25 & 60.60 & 18.18 & 21.21 & 57.57 & 21.21 & 21.21 & 69.69 & 15.15 & 15.15 & 0.6220\\ \hline 
    \end{tabular}}
\end{table}

\subsection{Inter-annotator Agreement} \label{s:ianoag}
\par In this Subsection, we check the consistency among the three annotators through the inter-annotator agreement for summary generation. We check this agreement on the two different types: 1) category assignment and 2) ground-truth summary. We discuss each of the measures next.  

\subsubsection*{\textbf{Category Assignment Inter-annotator Agreement}} \label{s:catint}
\par We compute Fleiss kappa~\cite{fleiss2013statistical} co-efficient score to calculate the inter-annotator agreement on the basis of the category assigned by an annotator given a tweet. We utilize the categories identified by all three annotators during ground-truth preparation for a disaster discussed in Section~\ref{s:dataprep}. Fleiss kappa provides an understanding of the similarities among the annotators. Our results, as shown in Table~\ref{table:Kappa} indicate that the Fleiss kappa co-efficient score ranges between $72-80\%$. We also found that the score is highly substantial, within the range between $61-80\%$~\cite{landis1977measurement}. Therefore, the kappa values across the category annotations ensure consistency across the annotators. 

\begin{table}[!htbp]
    \caption{Table shows the classification annotation agreement scores for $D_1$-$D_8$ datasets over the three annotators category annotated data.}
    \label{table:Kappa}
    \centering
    \resizebox{0.6\textwidth}{!}{\begin{tabular} {cc||cc}
        \hline
        {\bf Dataset} & {\bf Kappa score} & {\bf Dataset} & {\bf Kappa score} \\\hline 
        
        $D_1$   &  72.13 & $D_5$&  72.29 \\ 
        $D_2$   &  74.76 & $D_6$&  79.46 \\ 
        $D_3$   &  78.33 & $D_7$&  77.89\\ 
        $D_4$   &  75.21 & $D_8$&  78.12\\ \hline   
    \end{tabular}}
\end{table}

\subsubsection*{\textbf{Ground-truth Summary Inter-annotator Agreement}} \label{s:groundint}
\par To assess the consistency among the three annotators for summary creation, we measure it through summary-level Inter-annotator Agreement, where we compare their annotated summaries based on \textit{content similarity}, \textit{topical similarity}, and \textit{semantic similarity}, as discussed below.
 
\par \textbf{Content similarity:} We calculate \textit{syntactic similarity}, i.e., the distance between the tweet keywords based on their meaning~\cite{little2020semantic}. To calculate the \textit{syntactic similarity} between a pair of annotated summaries, we use Cosine Similarity~\cite{nguyen2010cosine}, $CosSIM(A_i, A_j)$ between the summary generated by a pair of annotators, $A_i$ and $A_j$ as: 

\begin{align}
   CosSIM(A_i, A_j) = \frac{ \vert Kw(A_i) \cap Kw(A_j) \vert}{\sqrt{\vert Kw(A_i) \vert \ \vert Kw(A_j) \vert}}
    \label{eq:consim}
\end{align}

where $Kw(A_i)$ and $Kw(A_j)$ are the keywords of $A_i$ and $A_j$ summaries, respectively. We consider only nouns, verbs and adjectives as keywords~\cite{khan2013multi}. Our observations as shown in Table~\ref{table:Sim} indicate that $CosSIM(A_i, A_j)$ ranges between $72-96\%$ which indicates that summaries written by different annotators are syntactically similar to each other.

\par \textbf{Topical similarity:}  
To calculate \textit{topical similarity}, $TopSIM(A_i, A_j)$ between any pair of annotators summaries, $A_i$ and $A_j$, we initially identify the top $15$ topics words present in an annotator summary using Latent Dirichlet Allocation (LDA)~\cite{blei2003latent}. We calculate the embedding for the topics of an annotator summary as the average of the values of the topic word vector. We consider the embedding of each topic word provided by Word2Vec~\cite{imran2016twitter}, which was trained on $52$M disaster-related messages from various types of disasters. We, finally, compute \textit{topical similarity} as Cosine Similarity of the topic vector embedding of $A_i$ and $A_j$ as: 

\begin{align}
   TopSIM(A_i, A_j) = \frac{\vec{T_i} \cdot \vec{T_j}}{\vert \vec{T_i} \vert \ \vert \vec{T_j} \vert}
    \label{eq:topsim}
\end{align}

where, $\vec{T_i}$ and $\vec{T_j}$ are the topic embedding vectors of $A_i$ and $A_j$ respectively. Our observations, as shown in Table~\ref{table:Sim} indicate that $TopSIM(A_i, A_j)$ ranges between $85-98\%$, which indicates that the summaries written by different annotators are topically similar to each other.

\begin{table*}
    \caption{Table shows Content Similarity, Topical Similarity, Semantic Similarity (using Word2Vec), and Semantic Similarity (using BERT) \% for $D_1$-$D_8$ datasets over the three annotators annotated summaries.}
    \label{table:Sim}
    \centering
    \resizebox{0.9\textwidth}{!}{\begin{tabular} {cccccccccc}
        \hline
        {\bf Dataset} & {\bf Annotator} & \multicolumn{2}{c}{\textbf{Content Similarity}} & \multicolumn{2}{c}{\textbf{Topical Similarity}} & \multicolumn{2}{c}{\textbf{Semantic Similarity}} & \multicolumn{2}{c}{\textbf{Semantic Similarity}} \\
        
        & & \multicolumn{2}{c}{\textbf{}} & \multicolumn{2}{c}{\textbf{}} & \multicolumn{2}{c}{\textbf{(using Word2Vec)}} & \multicolumn{2}{c}{\textbf{(using BERT)}} \\ \cline{3-10}

        & & {\bf $\bf A_2$} & {\bf $\bf A_3$} & {\bf $\bf A_2$} & {\bf $\bf A_3$} & {\bf $\bf A_2$} & {\bf $\bf A_3$} & {\bf $\bf A_2$} & {\bf $\bf A_3$} \\\hline 
        
        $D_1$   & $A_1$ & 83.85 & 81.93 & 87.06 & 87.49 & 99.12 & 98.94 & 99.58 & 99.26 \\ 
        $D_2$   & $A_1$ & 89.27 & 91.55 & 84.63 & 90.45 & 98.93 & 98.98 & 99.64 & 99.54 \\ 
        $D_3$   & $A_1$ & 89.64 & 90.82 & 91.36 & 85.32 & 99.03 & 98.95 & 99.41 & 99.69 \\ 
        $D_4$   & $A_1$ & 83.17 & 78.68 & 91.08 & 86.71 & 99.38 & 98.93 & 99.43 & 99.21 \\ 
        $D_5$& $A_1$ & 69.93 & 71.84 & 87.61 & 84.53 & 99.44 & 99.11 & 99.40 & 99.52 \\ 
        $D_6$& $A_1$ & 88.34 & 86.49 & 93.55 & 93.55 & 99.73 & 99.68 & 99.72 & 99.70 \\ 
        $D_7$& $A_1$ & 91.14 & 95.59 & 97.66 & 97.95 & 99.71 & 99.79 & 99.87 & 99.93 \\ 
        $D_8$& $A_1$ & 85.12 & 91.35 & 97.41 & 98.49 & 99.45 & 99.69 & 99.55 & 99.72 \\ \hline
    \end{tabular}}
\end{table*}

\par \textbf{Semantic similarity:} In order to understand the context similarity between the annotated summaries, we calculate \textit{semantic similarity}. \textit{Semantic similarity} measures the distance between the semantic meanings or semantic content of a pair of keywords~\cite{miller1991contextual}. We compute \textit{semantic similarity} using two different mechanisms, Word2Vec~\cite{imran2016twitter} and BERT~\cite{liu2019text}. We calculate $SemSIM_{W2V}(A_i, A_j)$ as the \textit{semantic similarity }score using Word2Vec between the vector embedding of the summary generated by a pair of annotators, $A_i$ and $A_j$ as: 

\begin{align}
   SemSIM_{W2V}(A_i, A_j) = \frac{\vec{W_i} \cdot \vec{W_j}}{\vert \vec{W_i} \vert \ \vert \vec{W_j} \vert}
    \label{eq:SemanticSim_W2V}
\end{align}

where, $\vec{W_i}$ and $\vec{W_j}$ are the embeddings of $A_i$ and $A_j$ respectively. We calculate $\vec{W_i}$ as the average of the values of the tweet embeddings. We calculate the embedding for each tweet as the average of the values of the word vector of the tweet. We consider the embedding of each word provided by Word2Vec~\cite{imran2016twitter}, which was trained on $52$M disaster-related messages from various types of disasters. We consider only nouns, verbs, and adjectives as words~\cite{khan2013multi}. As shown in Table~\ref{table:Sim}, our observation indicates that $SemSIM_{W2V}(A_i, A_j)$ ranges between $98-99\%$, which indicates that the summaries written by different annotators are highly semantic similar to each other. Similarly, we calculate $SemSIM_{BERT}(A_i, A_j)$ as the semantic similarity score using BERT between the vector embedding of the summary generated by a pair of annotators, $A_i$ and $A_j$ as: 

\begin{align}
   SemSIM_{BERT}(A_i, A_j) = \frac{\vec{B_i} \cdot \vec{B_j}}{\vert \vec{B_i} \vert \ \vert \vec{B_j} \vert}
    \label{eq:SemanticSim_BERT}
\end{align}

where, $\vec{B_i}$ and $\vec{B_j}$ are the embedding of $A_i$ and $A_j$ respectively. We calculate $\vec{B_i}$ as the average of the values of the tweet embedding. We consider the embedding for each tweet provided by a pre-trained language model known as BERT~\cite{liu2019text}. We use an uncased version of the BERT-base model with default hyperparameters. As shown in Table~\ref{table:Sim}, our observation indicates that $SemSIM_{BERT}(A_i, A_j)$ ranges above $99\%$, which indicates that the summaries written by different annotators are highly semantically similar to each other.

\begin{figure*}
    \centering
    \begin{subfigure}[b]{0.32\linewidth}
    \includegraphics[width=\linewidth]{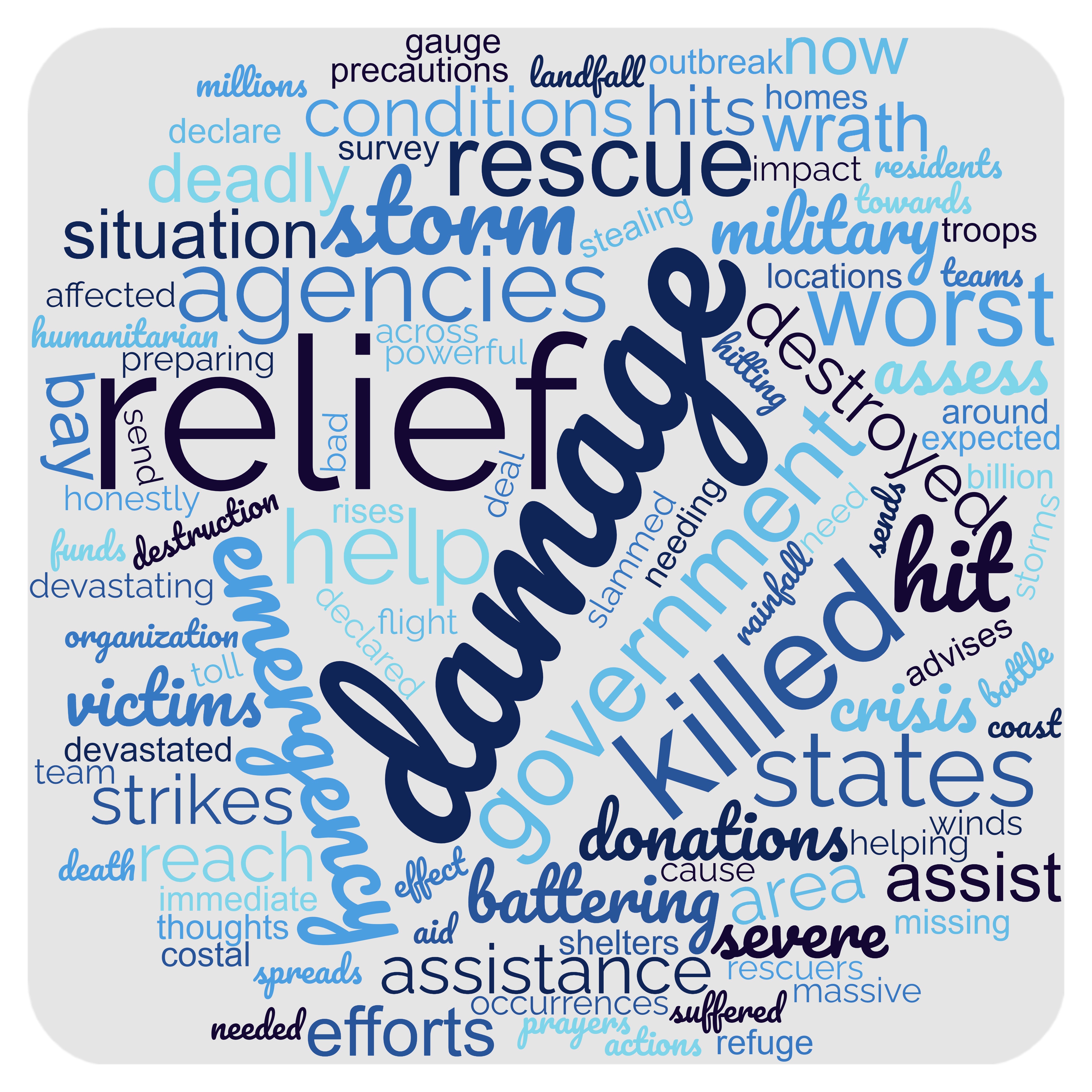}\caption{}
    \label{fig:A1_matthew}
    \end{subfigure}
    \begin{subfigure}[b]{0.32\linewidth}
    \includegraphics[width=\linewidth]{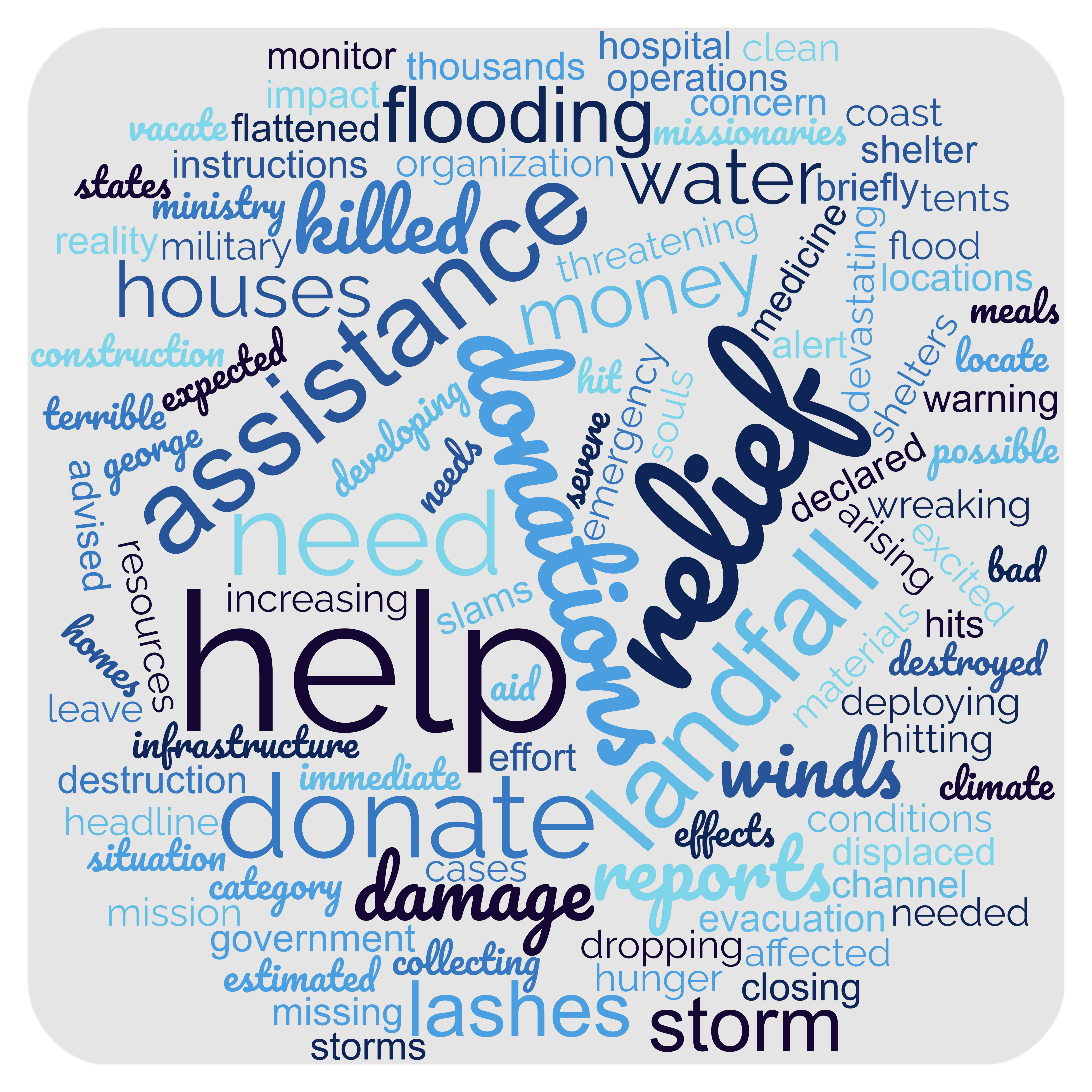}\caption{}
    \label{fig:A2_matthew}
    \end{subfigure}
    \begin{subfigure}[b]{0.32\linewidth}
    \includegraphics[width=\linewidth]{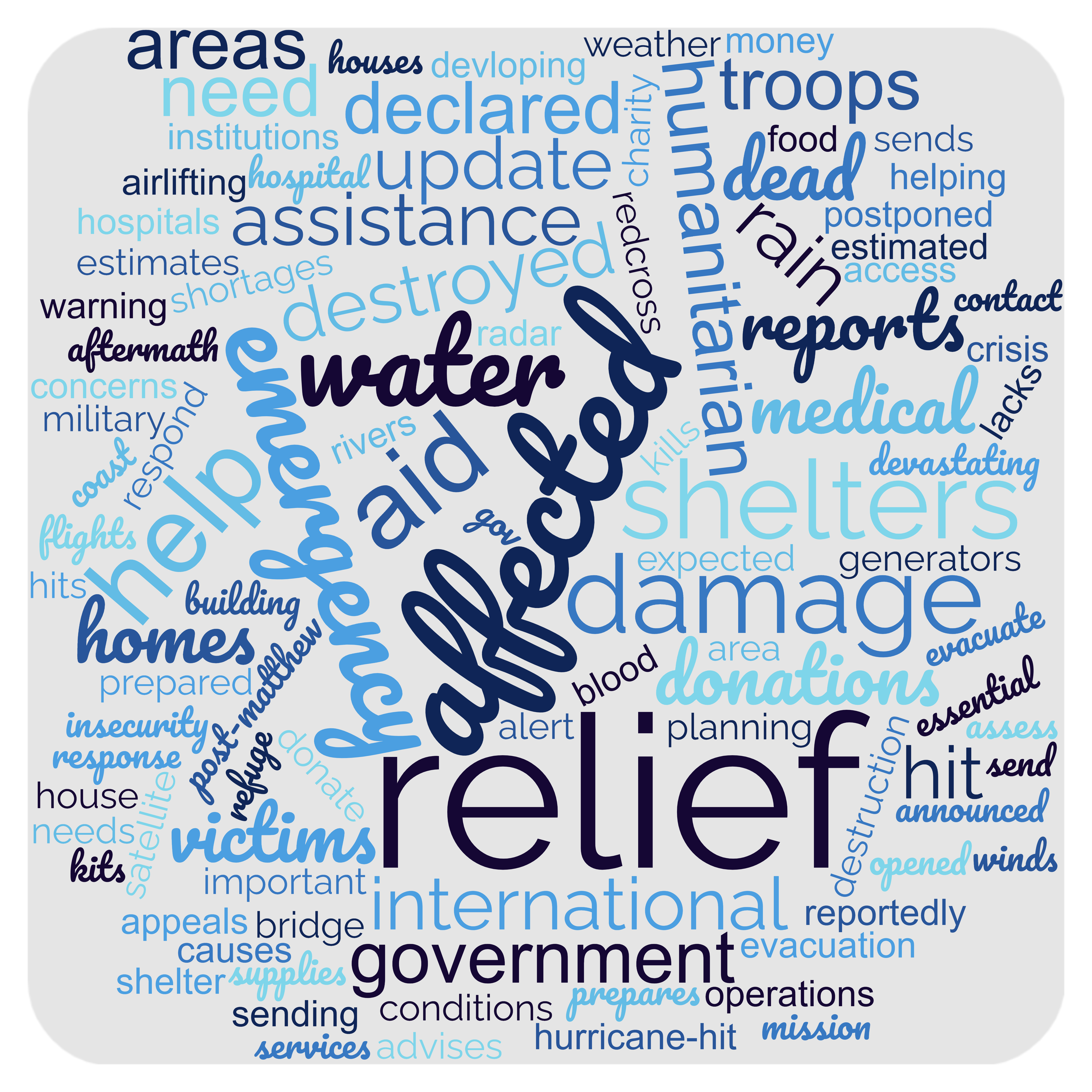}\caption{}
    \label{fig:A3_matthew}
    \end{subfigure}
    \begin{subfigure}[b]{0.32\linewidth}
    \includegraphics[width=\linewidth]{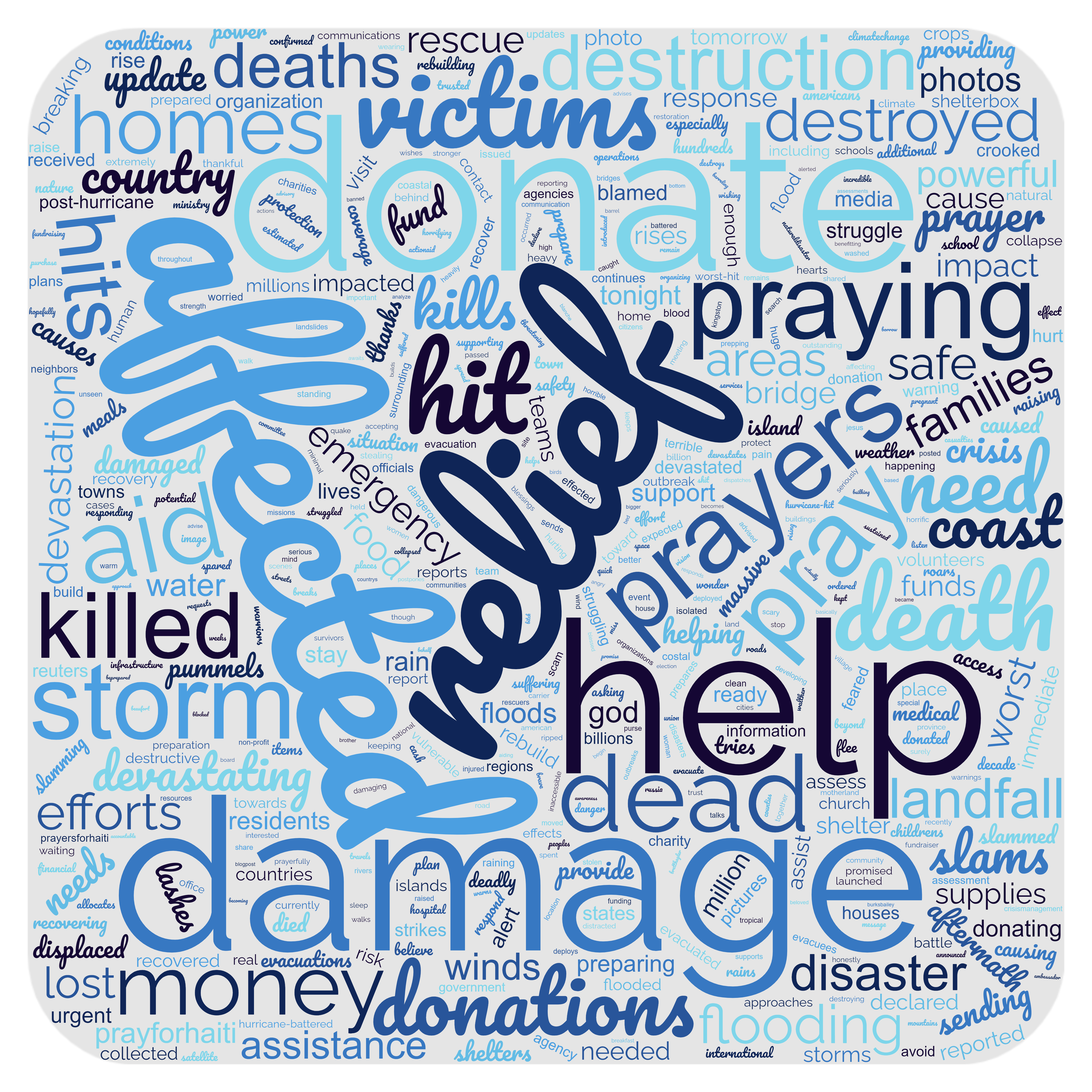}\caption{}
    \label{fig:matthew}
    \end{subfigure}\begin{subfigure}[b]{0.32\linewidth}
    \includegraphics[width=\linewidth]{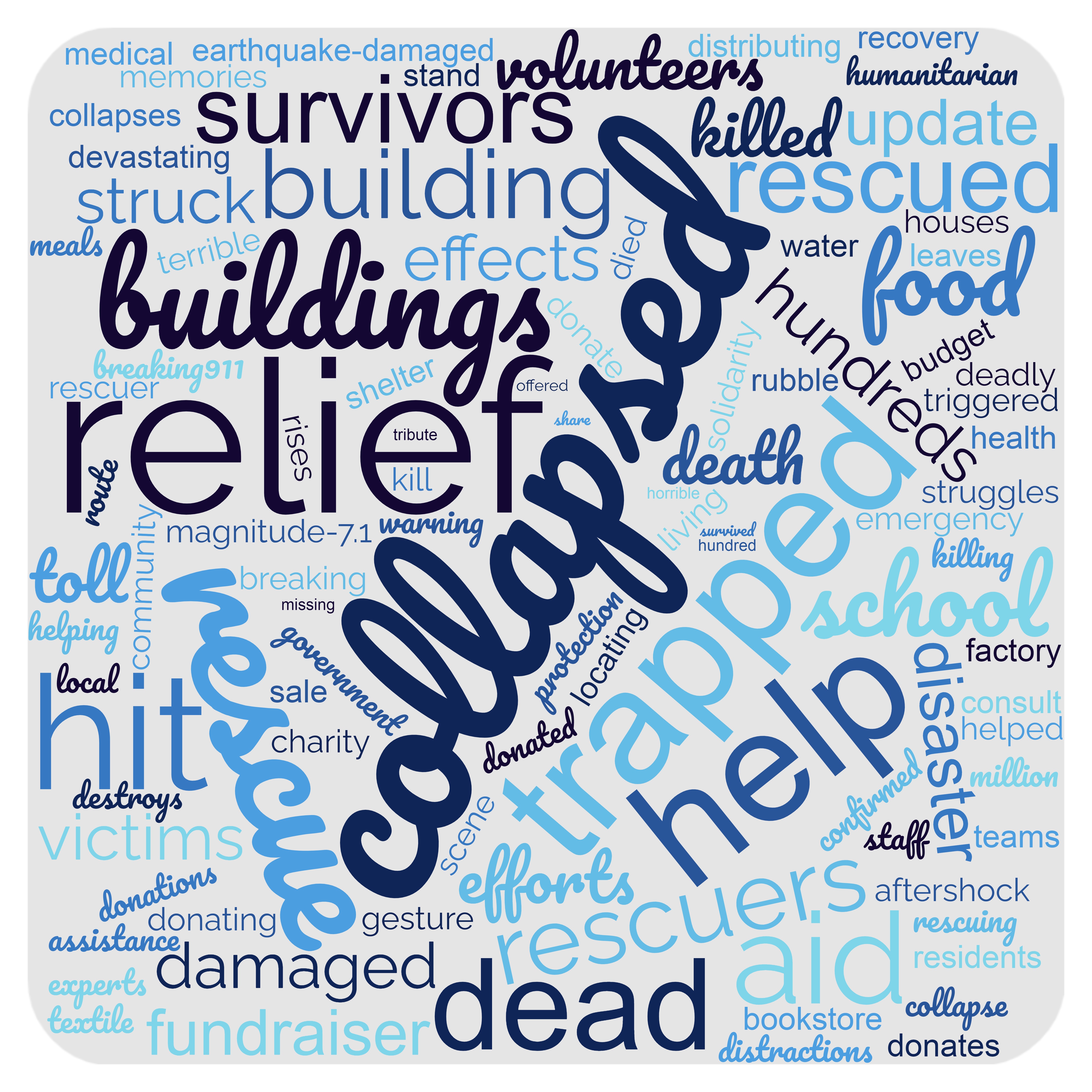}\caption{}
    \label{fig:A1_mexico}
    \end{subfigure}
    \begin{subfigure}[b]{0.32\linewidth}
    \includegraphics[width=\linewidth]{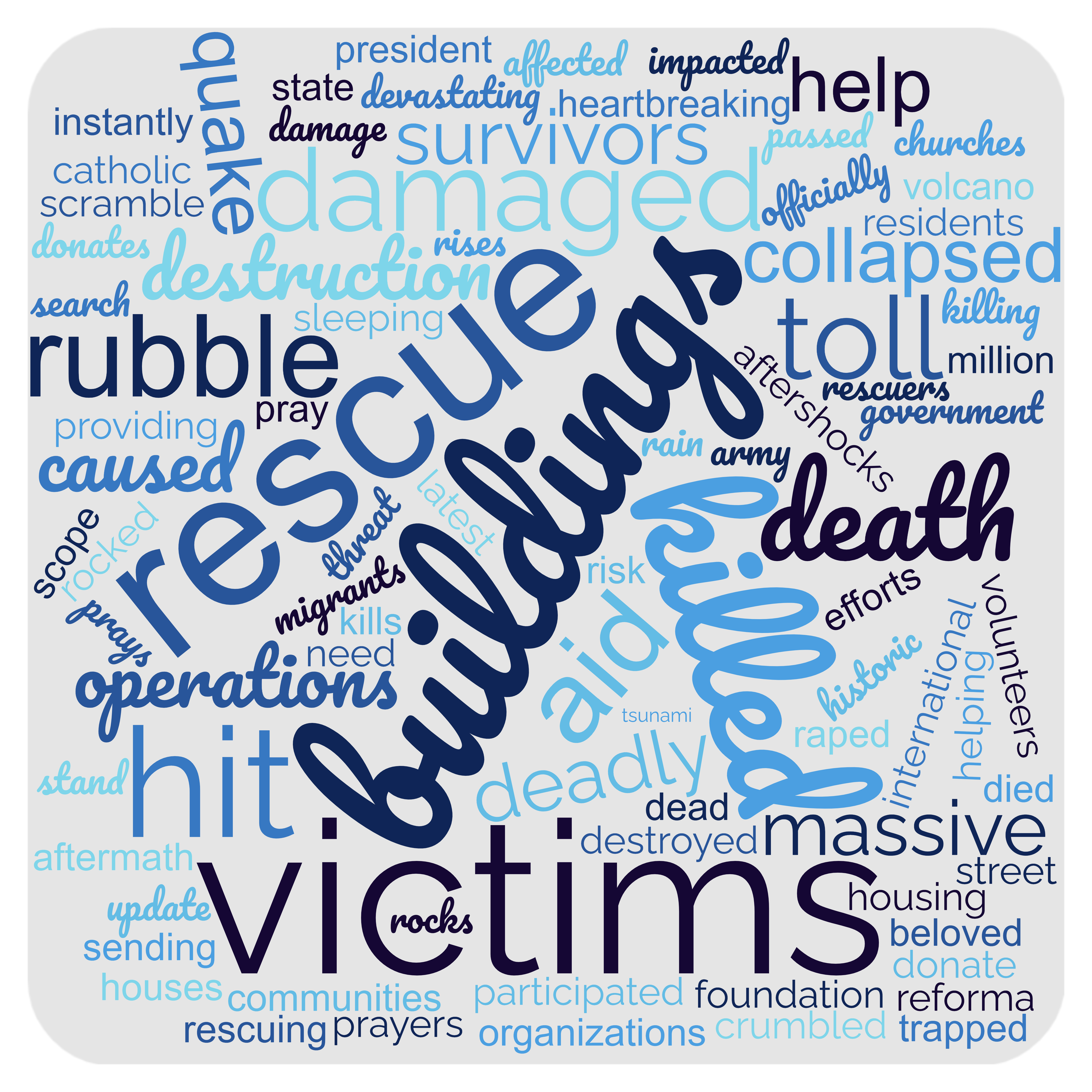}\caption{}
    \label{fig:A2_mexico}
    \end{subfigure}
    \begin{subfigure}[b]{0.32\linewidth}
    \includegraphics[width=\linewidth]{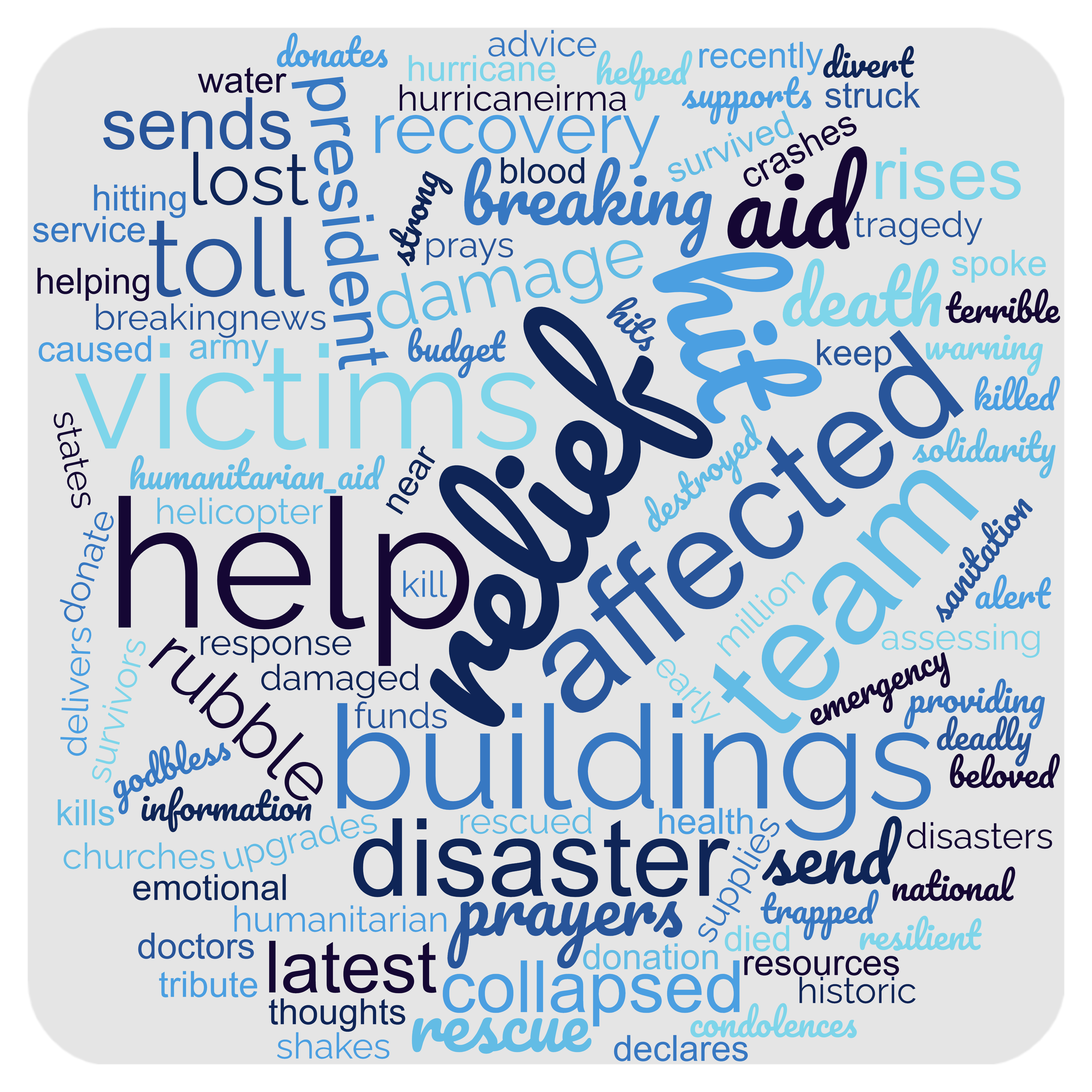}\caption{}
    \label{fig:A3_mexico}
    \end{subfigure}
    \begin{subfigure}[b]{0.32\linewidth}
    \includegraphics[width=\linewidth]{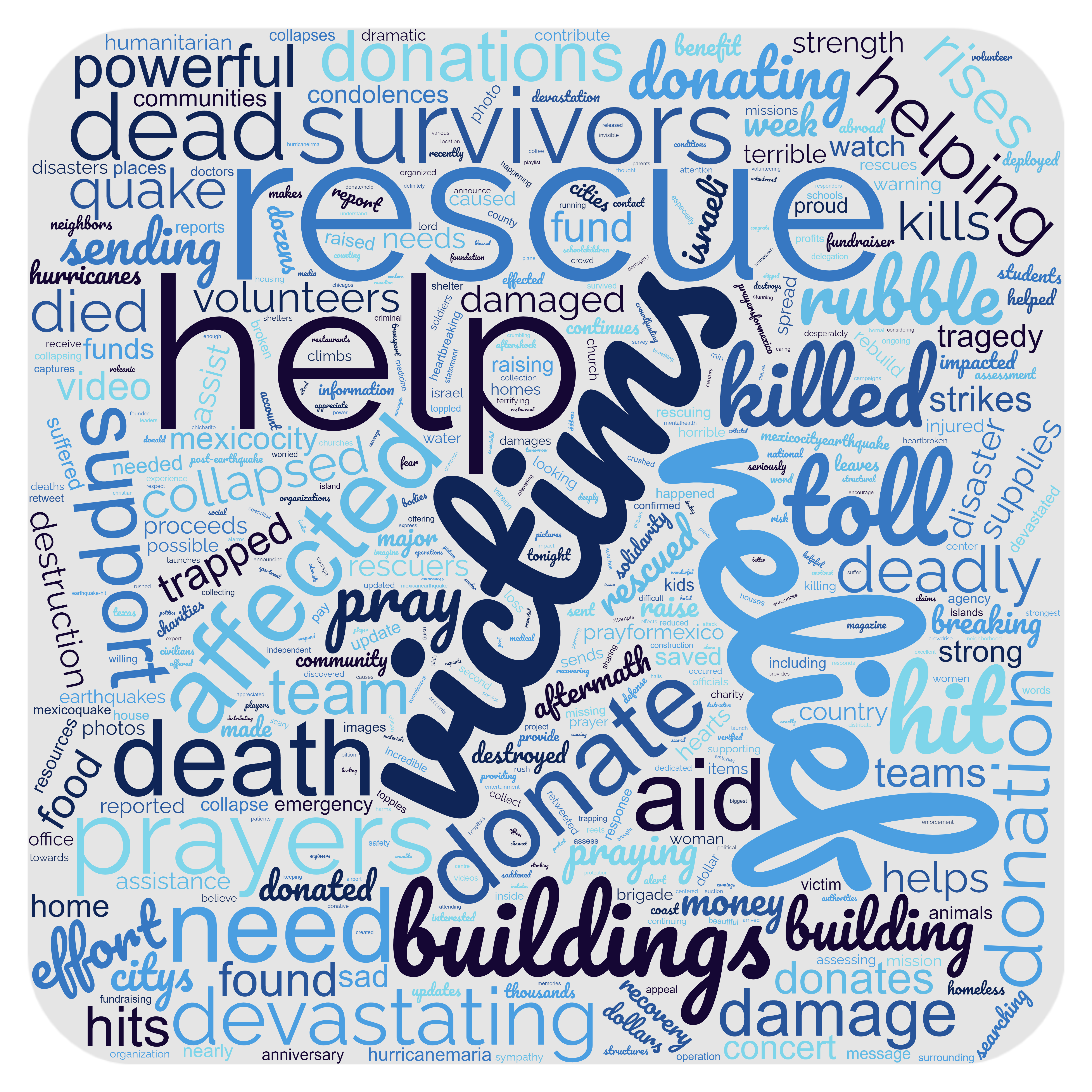}\caption{}
    \label{fig:mexico}
    \end{subfigure}
    \caption{Figure shows the word cloud of annotated summaries of $3$ annotators and the original tweet set for $D_1$ and $D_2$ disasters, such as $A_1$ of $D_1$ in Figure~\ref{fig:A1_matthew}, $A_2$ of $D_1$ in Figure~\ref{fig:A2_matthew}, $A_3$ of $D_1$ in Figure~\ref{fig:A3_matthew}, tweet set of $D_1$ in Figure~\ref{fig:matthew}, $A_1$ of $D_2$ in Figure~\ref{fig:A1_mexico}, $A_2$ of $D_2$ in Figure~\ref{fig:A2_mexico}, $A_3$ of $D_2$ in Figure~\ref{fig:A3_mexico}, and tweet set of $D_2$ in Figure~\ref{fig:mexico}.} 
    \label{fig:Wordcloud}
\end{figure*}

\subsection{Visualization of the Annotated Summaries}\label{s:visual} 
\par To visualize the content of the summaries provided by three annotators for a dataset, we use Word Cloud\footnote{\url{https://en.wikipedia.org/wiki/Tag\_cloud}} based representation of the tweets in the summary~\cite{castella2014word}. For a tweet summary, the word cloud represents only the most frequent words in the summary with the font size being directly proportional to the word frequency. We generate the word cloud of the summary generated by each annotator to understand whether the most frequent words are similar across annotators. We show Word Clouds of the summaries of $A_1$, $A_2$, and $A_3$ with original tweet set for $D_1$ and $D_2$ in Figure~\ref{fig:Wordcloud}. Our observations indicate that the occurrence of the most frequent words is similar across annotators and the original tweet set. From these, we can say that the summaries generated by the annotators cover all the important aspects of the given disaster event.  

\subsection{Impact on Increasing Number of Datasets on Supervised Approaches}\label{s:impact}
\par In this Subsection, we evaluate the performance of a supervised approach when it is trained on an expanded set of datasets that includes our newly introduced datasets along with the available datasets. To show this, we consider two supervised disaster tweet summarization approaches, IKDSumm proposed by Garg et al.~\cite{garg2024ikdsumm} and TSSuBERT proposed by Dusart et al.~\cite{dusart2023tssubert}. We train IKDSumm and TSSuBERT in $2$ different ways: 1) train with available disaster datasets (shown in Table~\ref{table:existdata}) and 2) an expanded set of datasets that includes our newly introduced dataset with the available disaster datasets. Then, we evaluate the performance of the summary generated by both ways in terms of the ROUGE-N, i.e., N=1, 2, and L, F1-score. For our experiment, we randomly select four different disaster datasets: $D_1$, $D_3$, $D_4$, and $D_6$. Our results, as shown in Table~\ref{table:impact}, indicate that IKDSumm performs better by $8.33-10.71$\%, $19.04-27.77$\%, $12-16.67$\% for ROUGE-1, ROUGE-2, and ROUGE-L, respectively, when it is trained on an expanded set of datasets than the available datasets. Similarly, TSSuBERT performs better by $8-9.43$\%, $16-25$\%, and $13.33-16.67$\% for ROUGE-1, ROUGE-2, and ROUGE-L, respectively, when it is trained on an expanded set of datasets than the available datasets. Therefore, based on the above experiment, we show our newly introduced datasets improve the performance of the supervised approaches. 

\begin{table} [!ht]
    \centering 
     \caption{Table shows F1-score of ROUGE-1, 2, and L of the summary generated by IKDSumm and TSSuBERT trained on an expanded set of datasets that includes our newly introduced datasets with the available datasets (i.e., IKDSumm-ExpData and TSSuBERT-ExpData) and trained on available disaster datasets (i.e., IKDSumm-AvlData and TSSuBERT-AvlData) for four disasters: $D_1$, $D_3$, $D_4$, and $D_6$.}
    \label{table:impact}
    \resizebox{0.7\textwidth}{!}{\begin{tabular}{clccc} \hline
        \textbf{Dataset} & \textbf{Approach} & \textbf{ROUGE-1} & \textbf{ROUGE-2} & \textbf{ROUGE-L}\\ \hline

                    & IKDSumm-ExpData    & \bf0.60 & \bf0.26 & \bf0.32 \\ 
        ${D_1}$     & IKDSumm-AvlData    & 0.55 & 0.21 & 0.27 \\ \hline
                    
                    & IKDSumm-ExpData    & \bf0.54 & \bf0.21 & \bf0.30 \\ 
        ${D_3}$     & IKDSumm-AvlData    & 0.49 & 0.17 & 0.25 \\ \hline
        
                    & IKDSumm-ExpData    & \bf0.57 & \bf0.18 & \bf0.25 \\ 
        ${D_4}$     & IKDSumm-AvlData    & 0.52 & 0.14 & 0.22 \\ \hline
        
                    & IKDSumm-ExpData    & \bf0.56 & \bf0.18 & \bf0.26 \\ 
        ${D_6}$     & IKDSumm-AvlData    & 0.50 & 0.13 & 0.22 \\ \hline
    \end{tabular}}
\bigskip

    \resizebox{0.7\textwidth}{!}{\begin{tabular}{clccc} \hline
        \textbf{Dataset} & \textbf{Approach} & \textbf{ROUGE-1} & \textbf{ROUGE-2} & \textbf{ROUGE-L}\\ \hline

                    & TSSuBERT-ExpData    & \bf0.57 & \bf0.25 & \bf0.30 \\ 
        ${D_1}$     & TSSuBERT-AvlData    & 0.52 & 0.21 & 0.26 \\ \hline
                    
                    & TSSuBERT-ExpData    & \bf0.53 & \bf0.17 & \bf0.27 \\ 
        ${D_3}$     & TSSuBERT-AvlData    & 0.48 & 0.13 & 0.23 \\ \hline
        
                    & TSSuBERT-ExpData    & \bf0.50 & \bf0.14 & \bf0.24 \\ 
        ${D_4}$     & TSSuBERT-AvlData    & 0.46 & 0.11 & 0.20 \\ \hline
        
                    & TSSuBERT-ExpData    & \bf0.49 & \bf0.16 & \bf0.25 \\ 
        ${D_6}$     & TSSuBERT-AvlData    & 0.45 & 0.12 & 0.21 \\ \hline
    \end{tabular} }
\end{table}

\subsection{Case Study: Utilization of the Datasets Features in Different Tasks}\label{s:utilifeat}
\par In this Subsection, we study how the additional features of the proposed datasets are helpful in the various NLP tasks. These dataset features are \textit{key-phrases}, \textit{category labels}, and \textit{relevance labels}. We already discussed how the \textit{relevance labels} are helpful for summary quality evaluation in Subsection~\ref{s:quantcomp}. We next discuss the utilization of \textit{category labels} in Subsection~\ref{s:catlabel} and the utilization of \textit{key-phrases} in Subsection~\ref{s:explain}.

\subsubsection{Utilization of Category Labels}\label{s:catlabel}
\par In this Subsection, we demonstrate how the \textit{category labels} provided with the proposed datasets can be useful in the various NLP tasks, like category identification of the tweet and determining the coverage quality of the summary. As we have already discussed how the \textit{category labels} are helpful for summary coverage quality evaluation in Subsection~\ref{s:qqeval}, we next discuss its role for category identification.

\subsubsection*{Identification of category of the tweets: }\label{s:catident}
\par The classification or categorization of disaster tweets is heavily used in literature for different purposes, such as category-specific summary generation~\cite{rudra2019summarizing, rudra2018identifying}, category-specific task assessment~\cite{priya2020taqe}, identifying the help request in a disaster~\cite{ullah2021rweetminer}, planing post-disaster relief strategies~\cite{rudra2017classifying} and so on. To evaluate the effectiveness of a tweet categorization approach, we require ground-truth \textit{category labels}. In the literature, we found both the supervised and unsupervised state-of-the-art approaches to identifying a category of a tweet. The supervised approach is BERT-based, whereas the unsupervised approach is graph-based and ontology-based. We select one prominent approach from each of the supervised and unsupervised approaches proposed by \cite{nguyen2022towards} and \cite{garg2023ontodsumm, dutta2019community}, respectively. For our experiment, we select $20$\% of the total tweets randomly for five disaster events, such as $D_1$, $D_3$, $D_5$, and $D_7$. We then compare the identified category of the tweets by the approaches proposed by~\cite{garg2023ontodsumm}, \cite{nguyen2022towards}, and \cite{dutta2019community} with the corresponding ground-truth \textit{category labels} in the proposed dataset in terms of F1-score, as shown in Table~\ref{table:catdet}. We observe that F1-scores are in the range of $0.984-0.997$, $0.7169-0.9077$, and $0.5330-0.7490$ for approaches used in~\cite{garg2023ontodsumm}, \cite{nguyen2022towards} and \cite{dutta2019community}. Therefore, based on the above experiment, we show the utilization of the \textit{category label} field of the proposed datasets and can say it is beneficial in the category identification task.

\begin{table} [!ht]
    \centering 
     \caption{Table shows F1-score of the category identification approaches proposed by~\cite{garg2023ontodsumm}, \cite{nguyen2022towards} and \cite{dutta2019community}, on comparing the ground-truth \textit{category label} annotations for five disasters: $D_1$, $D_3$, $D_5$, and $D_7$.}
    \label{table:catdet}
    \resizebox{0.5\textwidth}{!}{\begin{tabular}{clc} \hline
        \textbf{Dataset} & \textbf{Approach} & \textbf{F1-score} \\ \hline

                    & Approach used in~\cite{garg2023ontodsumm}     & \bf{0.9950} \\ 
        ${D_1}$     & Approach used in~\cite{nguyen2022towards}     & 0.8409 \\
                    & Approach used in~\cite{dutta2019community}    & 0.5330      \\ \hline
                    
                    & Approach used in~\cite{garg2023ontodsumm}     & \bf{0.9840} \\  
        ${D_3}$     & Approach used in~\cite{nguyen2022towards}     & 0.9077 \\
                    & Approach used in~\cite{dutta2019community}    & 0.5710      \\ \hline
        
                    & Approach used in~\cite{garg2023ontodsumm}     & \bf{0.9970} \\  
        ${D_5}$  & Approach used in~\cite{nguyen2022towards}     & 0.8395 \\
                    & Approach used in~\cite{dutta2019community}    &  0.7490     \\ \hline
        

                    & Approach used in~\cite{garg2023ontodsumm}     & \bf{0.9910} \\ 
        ${D_7}$  & Approach used in~\cite{nguyen2022towards}     & 0.7169 \\
                    & Approach used in~\cite{dutta2019community}    &  0.5840     \\ \hline
    \end{tabular} }
\end{table}

\subsubsection{Utilization of Key-phrases}\label{s:explain}
\par In this Subsection, we demonstrate the usefulness of \textit{key-phrases} provided with the proposed datasets in development of robust summarization algorithm, determination of the \textit{Diversity} in summary and the necessity of \textit{key-phrases} highlighting in summary. We discuss the role of \textit{key-phrases} for summary diversity quality evaluation in Subsection~\ref{s:qqeval}. We next discuss the utilization of \textit{key-phrases} for the remaining two tasks.

\subsubsection*{Development of a summarization algorithm:}\label{s:devlop}
\par Existing summarization approaches inherently perform two steps for creating a summary from input tweets: 1) ranking of the tweets based on the importance of tweets, and 2) selection of important tweets into the summary. For the ranking of the tweets, existing work proposed by Dusart et al.~\cite{dusart2023tssubert} integrate the tweet's word frequencies with the corresponding tweet embedding identified using the DistilBERT model to determine each tweet's importance score. With a huge training set, this approach can automatically find out important keywords and in turns can compute the importance score of tweet efficiently. However, with limited dataset this model fails to compute the importance score efficiently. Therefore, availability of \textit{key-phrases} can aid in determination of the tweet's importance even with a sparse disaster dataset. To empirically validate this, we consider two approaches, such as, Garg et al.~\cite{garg2024ikdsumm} and Dusart et al.~\cite{dusart2023tssubert} where \cite{garg2024ikdsumm} utilizes the \textit{key-phrases} to identify the tweet's importance and \cite{dusart2023tssubert} does not utilizes the \textit{key-phrases}. For our experiment, we randomly select $2$\% of the tweets from the $D_1$ dataset and compute the importance score of a tweet by \cite{garg2024ikdsumm} and \cite{dusart2023tssubert} approaches, respectively. We rank these tweets along with the importance score in descending order and request a meta annotator to score which ranked list comprises of more important tweets. However, the metaannotator has no knowledge of the underlying approaches for tweet importance selection and also, which list belongs to which approach. On the basis of the meta-annotator's decision, we observe that  \textit{key-phrases} helps to identify the importance of a tweet more efficiently.     

\par For our selection of tweets into the summary, we consider : 1) Importance of tweets from ranking and 2) \textit{Diversity} among the already selected tweets in the summary. To ensure \textit{Diversity} in summary tweets, we can utilize their \textit{key-phrases}. In order to evaluate the importance of \textit{key-phrases} in diversity calculation which effectively increases the summary quality, we follow two different variants of IKDSumm ~\cite{garg2024ikdsumm} by considering \textit{Diversity} and without \textit{Diversity} in tweet selection into the summary. While considering \textit{Diversity}, we iteratively select tweets with the maximum importance score and have minimum similarity (maximum diversity) with the already selected tweets into the summary. For \textit{Diversity} calculation, we utilize \textit{key-phrases} as shown in IKDSumm ~\cite{garg2024ikdsumm}. For the variant without considering \textit{Diversity}, we iteratively select tweets on the basis of only the importance score into the summary. For our experiment, we consider randomly selected four different disaster datasets, i.e., $D_1$, $D_3$, $D_5$, and $D_6$ to evaluate the performance of the summary generated by both the methods with the ground-truth summary using ROUGE-N F1-score. Our observations, as shown in Table~\ref{table:devlop}, indicate that the improvement in summary score of ROUGE-N F1-score ranges from $5.67-36.84$\% while we with considering \textit{key-phrases} for summary creation than without \textit{key-phrases}. Therefore, based on the above experiments, we show that our introduced \textit{key-phrases} filed with the proposed dataset is beneficial in developing a robust summarization approach for both the ranking as well as the selection of tweets into the summary.

\begin{table} [!ht]
    \centering 
     \caption{Table shows F1-score of ROUGE-1, 2 and L of the summary generated with considering \textit{Diversity} (IKDSumm-withDiv) and without considering \textit{Diversity} (IKDSumm-withoutDiv) by~\cite{dusart2023tssubert} on four disasters: $D_1$, $D_3$, $D_5$, and $D_6$.}
    \label{table:devlop}
    \resizebox{0.75\textwidth}{!}{\begin{tabular}{clccc} \hline
        \textbf{Dataset} & \textbf{Approach} & \textbf{ROUGE-1} & \textbf{ROUGE-2} & \textbf{ROUGE-L}\\ \hline

                    & IKDSumm-withDiv    & \bf0.59 & \bf0.25 & \bf0.32 \\ 
        ${D_1}$     & IKDSumm-withoutDiv    & 0.55 & 0.21 & 0.28 \\ \hline
                    
                    & IKDSumm-withDiv    & \bf0.53 & \bf0.20 & \bf0.29 \\ 
        ${D_3}$     & IKDSumm-withoutDiv    & 0.50 & 0.16 & 0.25 \\ \hline
        
                    & IKDSumm-withDiv    & \bf0.56 & \bf0.19 & \bf0.25 \\ 
        ${D_5}$     & IKDSumm-withoutDiv    & 0.50 & 0.12 & 0.21 \\ \hline
        
                    & IKDSumm-withDiv    & \bf0.57 & \bf0.18 & \bf0.26 \\ 
        ${D_6}$     & IKDSumm-withoutDiv    & 0.49 & 0.14 & 0.22 \\ \hline
    \end{tabular} }
\end{table}

\subsubsection*{Importance of key-phrases in understanding summary:} \label{s:nece}
\par Rudra et al.~\cite{rudra2018identifying} discusses that \textit{key-phrases} makes a summary more interpretable for users and further, provide the reasoning behind the selection of a particular tweet in the summary. In order to understand the necessity of \textit{key-phrases} in summary, we conduct an experiment where we provide two summaries 1) without \textit{key-phrases} and 2) with \textit{key-phrases} of each tweet to $5$ meta-annotators and ask them to select the best summary out of these two summaries. For this experiment, we utilize \textit{IKDSumm} summarization approach~\cite{garg2024ikdsumm} to generate the summary and datasets as $D_1$, $D_4$, and $D_7$.  Our observations, as shown in Table~\ref{table:ness} indicate that a huge majority of meta-annotators ($D_1$ - $80$\%, $D_4$ - $100$\%, $D_7$ - $100$\%) found highlighting \textit{key-phrases} provide better explanations for inclusion of a tweet into the summary compared to the other. Therefore, the identified \textit{key-phrases} field with the proposed datasets will certainly help the research community as it provides additional important information about the summary.    

\begin{table}[ht!]
    \caption{Table shows results of meta-annotators preference in percentage (AnnoPrefPer) of the summary generated by IKDSumm with highlighted \textit{key-phrases} and without highlighted \textit{key-phrases} for $D_1$, $D_4$, and $D_7$ datasets over $3$ annotators.}
    \label{table:ness}
    \centering
    \resizebox{0.6\textwidth}{!}{\begin{tabular} {clc}
        \toprule
        {\bf Dataset}   & {\bf Approach} & {\bf AnnoPrefPer} \\ \midrule
        $D_1$   & IKDSumm-withHighlight    & 80.00\\
                & IKDSumm-withoutHighlight & 20.00\\ \midrule
        $D_4$   & IKDSumm-withHighlight    & 100\\
                & IKDSumm-withoutHighlight & NA\\ \midrule   
        $D_7$   & IKDSumm-withHighlight    & 100\\
                & IKDSumm-withoutHighlight & NA\\  \bottomrule
    \end{tabular}}
\end{table}

\subsection{Case Study: Evaluation of Existing Summarization Approaches}\label{s:exp}
\par In this Subsection, we initially discuss the various existing state-of-the-art summarization approaches, followed by a comparison of the annotated ground-truth summaries for $8$ disaster tweet datasets.

\subsubsection{Existing Summarization Approaches}\label{s:baseline}
\par We categorize existing summarization approaches into \textit{content-based}, \textit{graph-based}, \textit{matrix factorization-based}, \textit{entropy-based}, \textit{semantic similarity-based} and \textit{deep learning-based} approaches. We select a few prominent tweet summarization approaches from each type which we discuss next. 

\begin{enumerate}
    
    \item \textit{Content-based Approaches:} We discuss the existing content-based summarization approaches, which are as follows:
        \begin{enumerate}
            \item \textit{LUHN}: Luhn et al.~\cite{luhn1958automatic} propose a frequency-based summarization approach that selects those tweets into a summary that has the highest frequency scoring words.

            \item \textit{SumBasic}: Nenkova et al.~\cite{nenkova2005impact} select those tweets into a summary that have the words with the maximum probability of occurrence.
            
            \item \textit{COWTS}: Rudra et al.~\cite{rudra2015extracting} select those tweets into a summary that has the maximum coverage of the content words (i.e., noun, main verb, and numerals).
    
            \item \textit{DEPSUB}: Rudra et al.~\cite{rudra2018identifying} propose a sub-events-based summarization approach that initially identifies the noun-verb pairs as the sub-events in each input tweet and then creates a summary by selecting the representative tweets by maximizing the information coverage of the disaster-specific keywords and the sub-events using Integer Linear Programming (ILP) based selection.
            
        \end{enumerate}

    \item \textit{Graph-based Approaches:} We discuss the existing graph-based summarization approaches, which are as follows:
        \begin{enumerate}
            \item \textit{Cluster Rank}: Garg et al.~\cite{garg2009clusterrank} initially identify the different clusters followed by utilizing the PageRank~\cite{page1999pagerank} algorithm to select tweets from each cluster in summary. 
            
            \item \textit{LexRank}: Erkan et al.~\cite{erkan2004lexrank} initially construct a sentence graph where nodes represent the sentences, and the edges are the content similarity between them and then determine Eigenvector~\cite{borgatti2005centrality} centrality score of each node in sentence graph. Finally, they create a summary by selecting the highest eigenvector centrality score into the summary.

            \item \textit{$EnSum$}: Dutta et al.~\cite{dutta2018ensemble} propose an ensemble graph-based tweet summarization approach, EnSum, which initially selects the most important tweets by using $9$ existing summarization algorithms. Then using these important tweets, they create a tweet similarity graph where the nodes represent the tweets, and the edges represent their similarity. They identify the different communities using a community detection algorithm and then create a summary by selecting the representative tweets from each community based on length, informativeness, and centrality scores. 
            
            \item \textit{COWEXABS}: Rudra et al.~\cite{rudra2019summarizing} propose a summarization framework initially identifying the most important tweets by maximizing the information coverage of the disaster-specific keywords in the extracted tweets. Then from these tweets, they create a graph where the nodes are disaster-specific keywords, and the edges represent the co-occurrence relationship between them. Finally, they select the tweets or tweet paths in summary, which can ensure maximum information coverage of the graph.
                       
            \item \textit{MEAD}: Radev et al.~\cite{radev2004mead} propose a centroid-based summarization approach where they initially segregate sentences using an agglomerative clustering approach. Further, they select the tweets from each cluster on the basis of the centrality score and diversity score in the summary.
        \end{enumerate}
        
    \item \textit{Matrix factorization-based Approaches:} We discuss the most popular matrix factorization-based summarization approaches in detail next. 

    \begin{enumerate}
        \item \textit{LSA}: Gong et al.~\cite{gong2001generic} propose a Latent Semantic Analysis (LSA) based approach, where they select the tweets which have the highest eigenvalue after Singular Value Decomposition (SVD) of the keyword matrix created from all the input tweets.
        
        \item \textit{SumDSDR}: He et al.~\cite{he2012document} propose a data reconstruction-based summarization approach, where they initially apply linear reconstruction and non-linear reconstruction objective functions to identify the relation between the sentences and then generate a summary by minimizing the reconstruction error.
    \end{enumerate}
    
    \item \textit{Entropy-based Approach:} Garg et al.~\cite{Garg2022Entropy} propose an entropy-based disaster tweet summarization approach, EnDSUM, that creates the summary by selecting the tweets which provide the maximum entropy and diversity in the summary.
    
    \item \textit{Semantic Similarity based Approach:}  Garg et al.~\cite{garg2023ontodsumm} propose a disaster-specific tweet summarization framework, OntoDSumm, which initially determines each tweet category using an ontology-based pseudo-relevance feedback approach followed by identification of the importance of each category which represents the number of tweets to be in summary from a category. Finally, they create a summary by selecting the representative tweets from each category based on the DMMR-based approach. 
    
    \item \textit{Deep learning-based Approaches:}  We discuss the existing deep learning-based summarization approaches, which are as follows:
    
    \begin{enumerate}
        \item \textit{TSSuBERT: } Dusart et al.~\cite{dusart2023tssubert} propose a summarization framework, TSSuBERT, which integrates the context of the tweets using the whole vocabulary related to the event to determine the tweet importance. Finally, they iteratively select the higher important tweets in summary. 
        
        \item \textit{GCNSUM: } Li et al.~\cite{li2021twitter} propose a GCN-based summarization framework that initially creates a tweet-similarity graph where the nodes are tweets and edges represent their content similarity. Then, they generate tweet hidden features for each tweet by applying GCN on the tweet similarity graph. Further, they determine the importance score of each tweet by combining the tweet's hidden features and the whole event embedding. Finally, they create a summary by iteratively selecting the most important tweets in the summary.

        \item \textit{IKDSumm: } Garg et al.~\cite{garg2024ikdsumm} propose a summarization framework, IKDSumm, which integrates the disaster-specific key-phrase identified using domain knowledge of ontology with the tweet to determine the tweet importance. Finally, they iteratively select tweets with higher importance and maximum diversity with the tweets already selected in the summary. 
    \end{enumerate}
\end{enumerate}

\begin{table*}
    \caption{Table shows F1-score of ROUGE-1 (R-1), ROUGE-2 (R-2), and ROUGE-L (R-L) score of the summaries generated by various baselines for $D_1$-$D_4$ datasets.}
    \label{table:results1}
    \centering 
    \resizebox{0.85\textwidth}{!}{\begin{tabular}{ccccccccccccc}
        \hline
        
        \textbf{Approach} & \multicolumn{3}{c}{\textbf{$D_1$}} & \multicolumn{3}{c}{\textbf{$D_2$}} & \multicolumn{3}{c}{\textbf{$D_3$}} & \multicolumn{3}{c}{\textbf{$D_4$}} \\ \cline{2-13}
        
        & \textbf{R-1} & \textbf{R-2} & \textbf{R-L} & \textbf{R-1} & \textbf{R-2} & \textbf{R-L} & \textbf{R-1} & \textbf{R-2} & \textbf{R-L} & \textbf{R-1} & \textbf{R-2} & \textbf{R-L}\\ \hline
        
        $Cluster Rank$  & 0.44 & 0.15 & 0.22 &      0.47 & 0.14 & 0.22 &    0.45 & 0.12 & 0.22 &    0.47 & 0.14 & 0.21 \\ 
        $Lex Rank$      & 0.39 & 0.09 & 0.20 &      0.40 & 0.10 & 0.18 &    0.38 & 0.08 & 0.19 &    0.33 & 0.09 & 0.19 \\ 
        $LSA$           & 0.47 & 0.17 & 0.25 &      0.47 & 0.13 & 0.22 &    0.45 & 0.14 & 0.23 &    0.47 & 0.13 & 0.21 \\ 
        $LUHN$          & 0.45 & 0.16 & 0.23 &      0.46 & 0.12 & 0.21 &    0.45 & 0.13 & 0.23 &    0.46 & 0.14 & 0.21 \\ 
        $MEAD$          & 0.39 & 0.10 & 0.22 &      0.47 & 0.11 & 0.22 &    0.46 & 0.14 & 0.24 &    0.42 & 0.07 & 0.17 \\ 
        $SumBasic$      & 0.42 & 0.13 & 0.22 &      0.46 & 0.13 & 0.21 &    0.41 & 0.09 & 0.21 &    0.43 & 0.09 & 0.20 \\ 
        $SumDSDR$       & 0.41 & 0.12 & 0.22 &      0.41 & 0.11 & 0.18 &    0.39 & 0.09 & 0.20 &    0.35 & 0.10 & 0.21 \\ 
        $COWTS$         & 0.43 & 0.14 & 0.23 &      0.46 & 0.12 & 0.22 &    0.45 & 0.13 & 0.22 &    0.44 & 0.11 & 0.21 \\ 
        $COWEXABS$      & 0.49 & 0.22 & 0.29 &      0.48 & 0.13 & 0.22 &    0.45 & 0.13 & 0.23 &    0.20 & 0.04 & 0.20 \\ 
        $DEPSUB$        & 0.52 & 0.21 & 0.23 &      0.44 & 0.12 & 0.22 &    0.44 & 0.14 & 0.23 &    0.45 & 0.11 & 0.21 \\ 
        $EnSum$         & 0.48 & 0.18 & 0.25 &      0.47 & 0.14 & 0.22 &    0.46 & 0.14 & 0.24 &    0.47 & 0.14 & 0.21 \\ 
        $EnDSUM$        & 0.55 & 0.21 & 0.27 &      0.52 & 0.17 & 0.24 &    0.52 & 0.14 & 0.26 &    0.51 & 0.16 & 0.24 \\ 
        $OntoDSumm$     & 0.57 & 0.24 & 0.30 &      0.51 & 0.17 & 0.26 &    0.52 & 0.19 & 0.27 &    0.54 & 0.17 & 0.24 \\ 
        $TSSuBERT$      & 0.55 & 0.23 & 0.28 &      0.50 & 0.15 & 0.25 &    0.51 & 0.16 & 0.26 &    0.48 & 0.12 & 0.21 \\ 
        $GCNSUM$        & 0.52 & 0.20 & 0.27 &      0.50 & 0.16 & 0.25 &    0.50 & 0.17 & 0.25 &    0.51 & 0.16 & 0.22 \\ 
        $IKDSumm$       & \bf0.60 & \bf0.26 & \bf0.32 &      \bf0.55 & \bf0.20 & \bf0.27 &    \bf0.54 & \bf0.21 & \bf0.30 &    \bf0.57 & \bf0.18 & \bf0.25 \\ \hline
    \end{tabular}}
\end{table*}

\begin{table*}
    \caption{Table shows F1-score of ROUGE-1 (R-1), ROUGE-2 (R-2), and ROUGE-L (R-L) score of the summaries generated by various baselines for $D_5$-$D_8$ datasets.}
    \label{table:results2}
    \centering 
    \resizebox{0.85\textwidth}{!}{\begin{tabular}{ccccccccccccc}
        \hline
        
        \textbf{Approach} & \multicolumn{3}{c}{\textbf{$D_5$}} & \multicolumn{3}{c}{\textbf{$D_6$}} & \multicolumn{3}{c}{\textbf{$D_7$}} & \multicolumn{3}{c}{\textbf{$D_8$}} \\ \cline{2-13}
        
        & \textbf{R-1} & \textbf{R-2} & \textbf{R-L} & \textbf{R-1} & \textbf{R-2} & \textbf{R-L} & \textbf{R-1} & \textbf{R-2} & \textbf{R-L} & \textbf{R-1} & \textbf{R-2} & \textbf{R-L}\\ \hline
        
        $Cluster Rank$      & 0.46 & 0.11 & 0.21    & 0.45 & 0.13 & 0.22 & 0.50 & 0.21 & 0.31 & 0.44 & 0.12 & 0.25 \\  
        $Lex Rank$          & 0.41 & 0.07 & 0.14    & 0.39 & 0.08 & 0.17 & 0.44 & 0.14 & 0.25 & 0.37 & 0.08 & 0.20 \\ 
        $LSA$               & 0.47 & 0.12 & 0.22    & 0.45 & 0.10 & 0.20 & 0.47 & 0.15 & 0.26 & 0.47 & 0.14 & 0.22 \\ 
        $LUHN$              & 0.44 & 0.08 & 0.18    & 0.48 & 0.12 & 0.21 & 0.49 & 0.17 & 0.29 & 0.48 & 0.15 & 0.23 \\ 
        $MEAD$              & 0.49 & 0.13 & 0.22    & 0.44 & 0.09 & 0.20 & 0.45 & 0.14 & 0.28 & 0.42 & 0.12 & 0.22 \\ 
        $SumBasic$          & 0.42 & 0.07 & 0.16    & 0.49 & 0.12 & 0.19 & 0.55 & 0.24 & 0.31 & 0.46 & 0.12 & 0.21 \\  
        $SumDSDR$           & 0.43 & 0.12 & 0.21    & 0.47 & 0.15 & 0.23 & 0.48 & 0.16 & 0.27 & 0.47 & 0.17 & 0.27 \\ 
        $COWTS$             & 0.43 & 0.09 & 0.19    & 0.50 & 0.16 & 0.20 & 0.46 & 0.17 & 0.26 & 0.47 & 0.14 & 0.21 \\ 
        $COWEXABS$          & 0.19 & 0.04 & 0.18    & 0.47 & 0.15 & 0.22 & 0.49 & 0.18 & 0.30 & 0.46 & 0.12 & 0.24 \\ 
        $DEPSUB$            & 0.50 & 0.12 & 0.22    & 0.43 & 0.11 & 0.19 & 0.49 & 0.17 & 0.28 & 0.44 & 0.10 & 0.20 \\  
        $EnSum$             & 0.48 & 0.10 & 0.20    & 0.42 & 0.09 & 0.20 & 0.49 & 0.17 & 0.29 & 0.47 & 0.13 & 0.24 \\ 
        $EnDSUM$            & 0.52 & 0.13 & 0.24    & 0.47 & 0.13 & 0.21 & 0.49 & 0.21 & 0.29 & 0.48 & 0.15 & 0.25 \\ 
        $OntoDSumm$         & 0.55 & 0.17 & 0.24    & 0.53 & 0.17 & 0.25 & 0.56 & 0.25 & 0.33 & 0.49 & 0.18 & 0.28 \\  
        $TSSuBERT$          & 0.50 & 0.12 & 0.21    & 0.47 & 0.15 & 0.23 & 0.50 & 0.18 & 0.29 & 0.40 & 0.12 & 0.22 \\  
        $GCNSUM$            & 0.46 & 0.10 & 0.20    & 0.46 & 0.14 & 0.19 & 0.51 & 0.21 & 0.31 & 0.42 & 0.13 & 0.22 \\  
        $IKDSumm$           & \bf0.56 & \bf0.19 & \bf0.25    & \bf0.56 & \bf0.18 & \bf0.26 & \bf0.57 & \bf0.25 & \bf0.34 & \bf0.52 & \bf0.20 & \bf0.29 \\ \hline
    \end{tabular}}
\end{table*}

\subsubsection{Comparison Results and Discussions}\label{s:result}
\par To evaluate the performance of the generated summaries of different baselines, we compare it with the ground-truth summaries using ROUGE-N~\cite{lin2004rouge} scores. The ROUGE-N score is a widely recognized metric in text summarization tasks which calculates the score by comparing the number of overlapping words between the system-generated and reference (or ground-truth) summaries. We use F1-score for $3$ different variants of the ROUGE-N score, i.e., N=$1$, $2$, and L, respectively. Our observations as shown in Table~\ref{table:results1} and~\ref{table:results2} indicate that IKDSumm ensures best ROUGE-N F1-score on $D_1-D_8$ datasets in comparison with different baselines followed by OntoDSumm. The reason for the high performance of IKDSumm is that it utilizes existing ontology knowledge instead of labelled training data to identify key-phrase and then utilizes these key-phrase to determine the tweet's importance to create a summary. Further, we observe that after IKDSumm, OntoDSumm ensures the best ROUGE-N F1-score for $D_1-D_8$, followed by TSSuBERT. The reason for the high performance of OntoDSumm is that it ensures each category's representation in a summary and handles the information diversity in summary tweets from each category. The performance of Lex Rank is the worst except for $D_4-D_5$, and COWEXABS is the worst for $D_4-D_5$ as both of these approaches do not ensure the category representation and information diversity in summary.

\section{Conclusions}\label{s:con}

In this study, we introduce a compilation of annotated datasets designed for the purpose of summarizing disaster-related tweets. Annotated ground-truth summary aids in enhancing the efficiency of supervised learning methods for both training and evaluation processes. The \textit{ADSumm} dataset comprises of tweets and corresponding ground-truth summaries for a total of eight distinct disaster events. These events encompass both man-made and natural disasters, originating from seven diverse geographical areas/country. In addition, this study also provides three supplementary components in conjunction with the datasets: \textit{category labels}, \textit{key-phrases}, and \textit{relevance labels}. The \textit{category label} is used to ensure that all categories are adequately represented in the summary. The \textit{relevance label} is employed to assess the quality of the summary. Lastly, \textit{key-phrases} are utilized to provide justification or explanation for the inclusion of a specific tweet in the summary. The inclusion of the recently incorporated datasets for training leads to a notable enhancement of $8-28$\% in the ROUGE-N F1-score for the supervised summarization methods. This study also presents a detailed methodology for preparing the ground-truth summary. Additionally, we discuss the practical applications of the supplementary features within the datasets for various Natural Language Processing (NLP) tasks. Moreover, we provide experimental results (quantitative as well as qualitative) to ensure the quality of the annotated ground-truth summary. In addition, we evaluate the effectiveness of different state-of-the-art summarization methods on these datasets by measuring their performance using the ROUGE-N F1-score. It is anticipated that the utilization of these datasets will contribute to the advancement of more resilient and effective catastrophe tweet summarizing methods. Consequently, this will facilitate the provision of assistance by humanitarian groups and government authorities.

\bibliography{sn-bibliography}%


\end{document}